\def\APResearchDesignOverride{1}
\documentclass[designerfonts=true]{allpurpose}

\usepackage{microtype}
\usepackage{hyperref}
\usepackage{url}
\usepackage{booktabs}
\usepackage{graphicx}
\usepackage{lineno}
\usepackage{amsmath}
\usepackage{booktabs}
\usepackage{pifont}

\usepackage{booktabs}
\usepackage{pifont}
\usepackage{xcolor}
\newcommand{\gck}{\textcolor{green!60!black}{\ding{51}}}

\usepackage{tcolorbox}
\usepackage{wrapfig}
\tcbset{
    colback=gray!5,
    colframe=black,
    boxrule=0.5pt,
    arc=2pt,
    left=6pt,
    right=6pt,
    top=6pt,
    bottom=6pt
}

\definecolor{darkblue}{rgb}{0, 0, 0.5}
\hypersetup{colorlinks=true, citecolor=darkblue, linkcolor=darkblue, urlcolor=darkblue, hypertexnames=false}
\usepackage{algorithm}
\usepackage{algpseudocode}
\usepackage{enumitem}
\usepackage{xcolor}
\usepackage{colortbl}
\newcommand{\rxmark}{\textcolor{red!70!black}{\ding{55}}}

\usepackage{tikz}
\usetikzlibrary{arrows.meta,positioning,calc}
\usepackage{pgfplots}
\pgfplotsset{compat=1.18}

\title[SoundnessBench]{SoundnessBench: Can Your AI Scientist Really Tell Good Research Ideas from Bad Ones?}
\paperstatus{Preprint}

\paperlink[globe]{Project}{https://hosytuyen.github.io/projects/SoundnessBench}
\paperlink[code]{Code repository}{https://github.com/hosytuyen/SoundnessBench}
\paperlink[database]{Datasets}{https://huggingface.co/datasets/hosytuyen/SoundnessBench}
\paperemail{stho@umd.edu}

\author{
  Sy-Tuyen Ho\affmark{1} \quad
  Minghui Liu\affmark{1}\quad
  Huy Nghiem\affmark{1}\quad
  Hieu Tran\affmark{1}\quad
  Aakriti Agrawal\affmark{1}\quad
  \apauthor{Yongkyun Lee}\quad
  \apauthor{Mingyu Kang}\quad
  \apauthor{Andrew Adams Schoen}\quad
  \apauthor[1]{Furong Huang}
}
\affiliations{\apaffiliation\qquad\affmark{1}University of Maryland}

\abstracttext{Autonomous AI research agents aim to speed up scientific discovery by automating the research pipeline from hypothesis generation to peer review. However, existing benchmarks rarely test a fundamental bottleneck whether LLMs can judge the methodological viability of a research idea $\textbf{before}$ investing time and computational resources. Without a reliable ``first-gate'' filter, autonomous agents risk scaling flawed methodology rather than accelerating meaningful science. We introduce $\textbf{SoundnessBench}$, a curated benchmark of 1,099 real research proposals, with source verified extraction, reviewer-derived proxy labels, and proposal-only expert judgments on a subset of 327 proposals. Across 12 frontier LLMs, our evaluation reveals a pervasive $\textbf{optimism bias}$: standard prompting yields a $74.0\%$ mean false-positive rate. Aggressive prompting reduces false approvals but substantially increases false rejections. Additional analyses show that optimism persists under controls for public corpus contamination, and surface features, as well as in literature-grounded agent evaluation. These results suggest a limitation in judging scientific soundness, indicating that current LLMs are not yet reliable as standalone gatekeepers for scientific rigor.}

\begin{document}

\maketitle

\section{Introduction}

Autonomous AI research agents are rapidly advancing in technical sophistication. Recent frameworks can autonomously generate hypotheses, implement code, execute experiments, and draft full manuscripts with minimal human oversight~\citep{lu2024aiscientist, yamada2025aiscientistv2, schmidgall2025agentlab, karpathy2026autoresearch}. While this progress promises a future of accelerated discovery, it also exposes a critical but underaddressed bottleneck: \textit{scientific triage}. Before commiting to extensice experiments, a researcher must decide whether a proposed idea and its experimental design are methodologically sound. In research, this ``first-gate'' judgment is an important defense against wasting months of labor and substantial computational resources.

\paragraph{The Soundness Gap.} Existing evaluations of LLM agents largely overlook this decision point. Current benchmarks~\citep{chan2024mle, starace2025paperbench, wu2025innovatorbench, lupidi2026airs} primarily emphasize reproducing existing results or deliver runnable artifacts. They do not test whether an agent can critically evaluate the viability of a proposal in the first place. This leaves a fundamental question unanswered: 

\begin{quote}
Can LLMs reliably reject flawed research designs before they are executed? 
\end{quote}

Without a robust upfront filter, autonomous agents do not necessarily accelerate science. Rather, they risk scaling ``bad science'' by automating the pursuit of unsound research designs.

\begin{table*}[ht]
\centering
\renewcommand{\arraystretch}{1.2}
\caption{\textbf{Comparison with related research-agent benchmarks.}
\textbf{Stage} indicates the research stage assessed (pre-execution,
execution, or post hoc). \textbf{Benchmark Task} describes the task,
and \textbf{Evaluation Input} describes the evidence available to
the evaluator. Among the benchmarks listed, SoundnessBench uniquely
combines \textbf{pre-execution evaluation},
\textbf{direct methodological-soundness judgment}, and
\textbf{proposal-only input}.}
\label{tab:benchmark_comparison}
\resizebox{\textwidth}{!}{%
\begin{tabular}{lcccccc}
\toprule
\textbf{Benchmark} & \textbf{Stage} & \textbf{Benchmark Task} & \textbf{Evaluation Input} 
& \textbf{Ground Truth} & \textbf{Soundness} & \textbf{Pre-Exec.} \\
 & & & & \textbf{Source} & \textbf{Focus} & \textbf{Judgment} \\
\midrule
MLE-Bench         
& Exec.     & Engineering   & Task desc. + Exp. Results      
& Kaggle outcome        & \rxmark{} & \rxmark{} \\

PaperBench      
& Exec.     & Replication   & Full paper + Exp. Results  
& Human rubric          & \rxmark{} & \rxmark{} \\

InnovatorBench      
& Exec.     & Research loop & Task desc. + Exp. Results    
& Outcome metric        & \rxmark{} & \rxmark{} \\

AIRS-Bench      
& Exec.     & Multi-task    & Task desc. + Exp. Results          
& Expert annotation     & \rxmark{} & \rxmark{} \\

ResearchGym
& Exec.     & Research loop & Task desc. + Exp. Results    
& Outcome metric        & \rxmark{} & \rxmark{} \\

\midrule

\cite{si2024llmideas} 
& Pre-exec.      & Novelty judge  & Proposal           
& Human study           & \rxmark{} & \gck{} \\

Hindsight               
& Post-hoc      & Impact pred.  & Idea          
& Future citation/venue        & \rxmark{} & \rxmark{} \\

RINoBench               
& Pre-exec.      & Novelty judge & Idea + related works            
& Expert label          & \rxmark{} & \gck{} \\

\cite{tong2026scientifictaste} 
& Post-hoc      & Impact pred. & Title + abstract        
& Future citation/venue      & \rxmark{} & \rxmark{} \\

\midrule

\textbf{SoundnessBench}                
& \textbf{Pre-exec.} & \textbf{Methodological soundness} 
& \textbf{Proposal} & \textbf{Expert label}  
& \gck{} & \gck{} \\

\bottomrule
\end{tabular}%
}
\end{table*}

In this work, we use \textbf{scientific soundness} in a narrowly defined sense: proposal-stage methodological integrity in ML research, not eventual impact, novelty, acceptance, or universal validity across all scientific domains. Given a hypothesis, the core question is whether the experimental design is capable of testing it rigorously. In the lifecycle of an autonomous agent, this serves as a \textit{pre-computation check}. Without this capability, agents risk implementing flawed experiments that appear structurally correct but cannot support their intended conclusions. By isolating proposal-stage evidence from downstream results, we test the model's ability to identify visible methodological flaws, such as improper baselines, data leakage, or mismatched metrics, before execution incurs high costs. However, the utility of such a check depends on its stability under changes in question framing. Prior work on sycophancy and prompt fragility suggests that LLM judgments can be sensitive to how a question is framed~\citep{sharma2024sycophancy, syntacticfragility, causalt5k}. If scientific evaluation inherits these vulnerabilities, autonomous AI agents may suffer from inconsistent filtering, alternating between reckless optimism and destructive skepticism based on minor prompt variations.

We introduce SoundnessBench, a large-scale benchmark for pre-execution scientific judgment in machine learning research. We processed $35,209$ initial ICLR submissions from 2022--2026 to identify a curated set of $1,099$ proposals. This covers 39 primary areas in machine learning research. For large-scale evaluation, we derive proxy labels from high-agreement ICLR reviewer soundness sub-scores rather than acceptance decisions or overall ratings. Because those reviewers saw full papers, we further complement the proxy labels with proposal-only judgments from 12 machine-learning researchers on a subset of 327 proposals. The reconstruction pipeline uses near-verbatim extraction, outcome masking, and retrieval-backed claim verification to keep each proposal faithful to its source paper while withholding downstream results.

Across 12 frontier LLMs, our evaluations reveal a pervasive \textbf{optimism bias}. Under standard prompting, models frequently classify low-soundness proposals as high-soundness, yielding a mean false-positive rate of 74.0\%. An aggressive prompt reverses the error balance rather than resolving it. The false-positive rate falls to 19.9\%, but high-soundness recall falls to 36.2\%. The pattern persists with proposal-only expert labels. On the 327 annotated proposals, standard prompting yields a mean of 64.8\% false-positive rate. These results suggest that current frontier models are not yet reliable as standalone first-gate critics for ML research proposals. Their judgments are inaccurate on many flawed designs and sensitive to prompt framing.

We further probe alternative explanations using a literature grounded agent, model scale and instruction tuning comparisons, a reduced contamination-risk split, surface-feature baselines, and controlled injections of methodological flaws. These analyses suggest that the optimism bias reflects a real weakness in current model judgment rather than a single dataset artifact.

\textbf{Our contributions are:}
\begin{itemize}[leftmargin=*,itemsep=0in, topsep=0in, parsep=0in]
    \item \textbf{A benchmark for pre-execution ML soundness:} SoundnessBench comprises 1,099 proposals spanning 39 ICLR primary areas, with reviewer-derived proxy labels and proposal-only judgments from 12 ML experts for a 327-proposal subset.
    \item \textbf{A source-grounded reconstruction pipeline:} Reviewer-agreement filtering, outcome masking, near-verbatim extraction, and retrieval-backed atomic-claim verification reduce label ambiguity, result leakage, and extraction drift.
    \item \textbf{An evaluation of first-gate model judgment:} Across 12 frontier LLMs and multiple label sources, we identify optimism under standard prompting and over-conservatism under aggressive prompting, and test the robustness of this pattern against literature grounding and a range of potential confounders.
\end{itemize}
\section{SoundnessBench: Benchmark Reconstruction} \label{sec:soundnessbench}

\textbf{Task definition.} SoundnessBench evaluates whether an LLM can assess the methodological soundness of a research proposal before observing experimental results. Given a hypothesis and experimental plan, the evaluator judges whether the proposed experiments can adequately test the hypothesis and support the intended conclusions. The task concerns methodological validity, rather than novelty, presentation quality, or whether the hypothesis will ultimately prove correct.

\textbf{Challenges and Design Choices.}
Constructing a proposal-only soundness benchmark from real submissions requires addressing four challenges.
First, \textit{label scope}: ICLR reviewers assessed full papers, while our models see results-masked proposals. We therefore use reviewer soundness sub-scores as proxy labels for large-scale evaluation and complement them with proposal-only expert annotations on a subset.
Second, \textit{label reliability}: both sources involve subjective judgments. We filter ICLR labels by reviewer agreement and exclude ambiguous middle-score cases. Expert annotations directly address the proposal-stage task on a subset. Agreement between the two sources provides an additional cross-check.
Third, \textit{information leakage}: completed papers contain results and acceptance cues that can shortcut first-gate assessment, so we exclude these and retain only proposal components.
Fourth, \textit{extraction fidelity}: extraction can introduce drift, so we preserve near-verbatim wording and check extracted claims against supporting passages in the original submission (Alg.~\ref{alg:verification}).

Our design choices combine proposal-only inputs, reviewer-derived proxy labels, direct expert annotations, and source-verified extraction to support evaluation of pre-execution scientific judgment.

\begin{figure}[t]
    \centering
    \includegraphics[width=\linewidth]{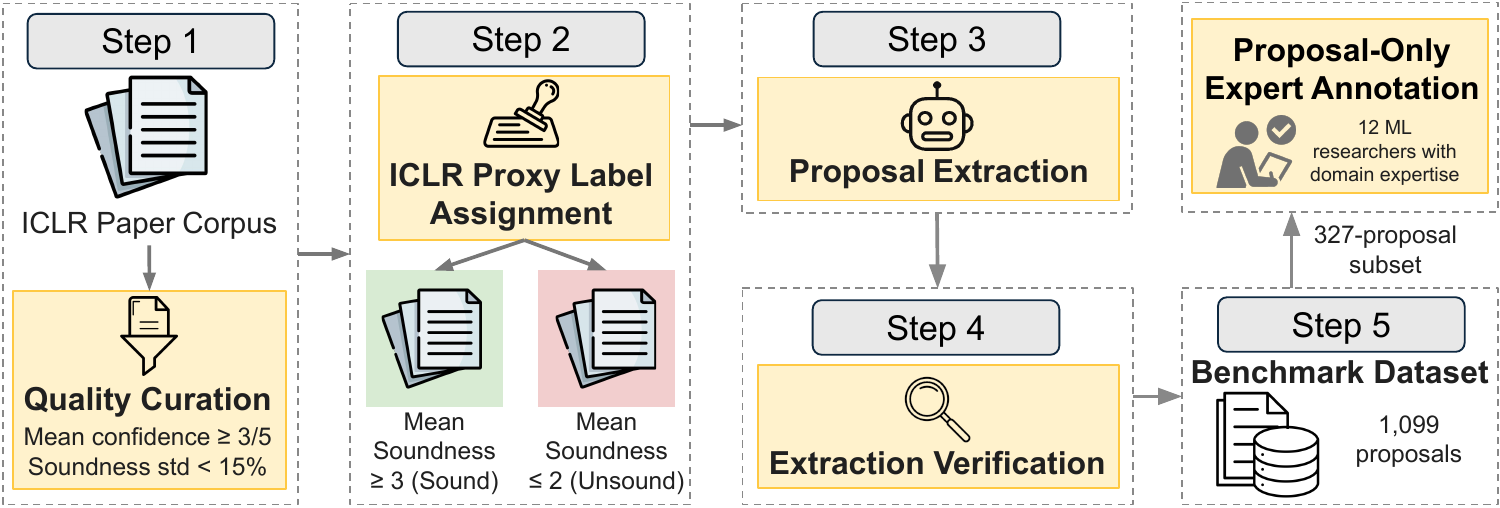}
    \caption{\textbf{SoundnessBench pipeline:} (1) collect ICLR papers with reviewer metadata and filter for high reviewer agreement; (2) derive high/low-soundness labels; (3) extract a near-verbatim research proposal without revealing experimental results; (4) audit extraction fidelity with retrieve-then-verify atomic claims; and (5) assemble the final benchmark.}\label{fig:pipeline}
\end{figure}

\begin{figure}[t]
    \centering
    \includegraphics[width=\linewidth]{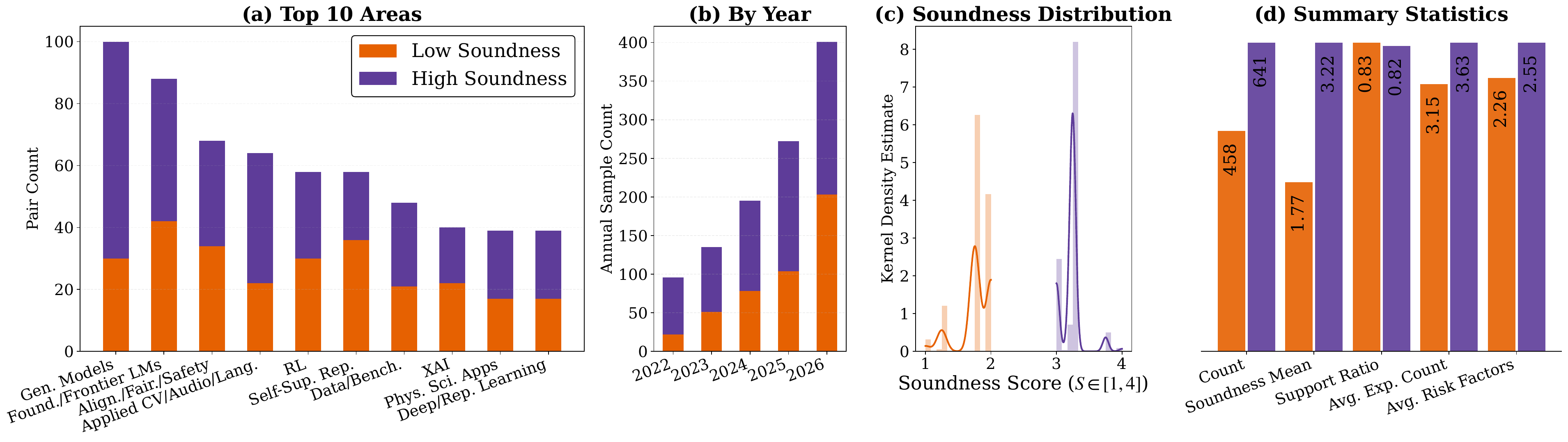}
    \caption{\textbf{SoundnessBench dataset statistics.} The benchmark contains 1,099 proposals, including 458 low-soundness and 641 high-soundness instances. \textbf{(a)} The subfield distribution across papers reflects the ICLR corpus composition. \textbf{(b)} Temporal coverage spans ICLR 2022--2026. \textbf{(c)} Soundness score density shows separation between low-soundness ($S \leq 2$, mean $= 1.77$) and high-soundness ($S \geq 3$, mean $= 3.22$) groups, supporting the chosen label boundary. \textbf{(d)} Low- and high-soundness pair-count statistics in SoundnessBench.}\label{fig:stats}
\end{figure}

\begin{algorithm}[ht]
\caption{Extraction Verification}\label{alg:verification}
\begin{algorithmic}[1]
\Require Source PDF $\mathcal{P}$, extracted pair $\hat{p} = (h, e)$,
         threshold $\tau$, chunk size $\ell$, chunk overlap $o$, retrieval depth $k$
\Ensure Fidelity score $\rho \in [0, 1]$, and binary pass/fail decision

\State $T \gets \textsc{ExtractText}(\mathcal{P})$
    \Comment{Read text from the source PDF}
\State $C \gets \textsc{Chunk}(T,\ell,o)$
    \Comment{Split text into overlapping passages}
\State $A \gets \textsc{DecomposeAtomicClaims}(\hat{p})$
    \Comment{Split into checkable claims}
\State $\mathrm{supported} \gets 0$
\For{each claim $c \in A$}
    \State $E \gets \mathrm{BM25}(c,C,\text{top-}k=k)$
        \Comment{Retrieve the $k$ most relevant passages}
    \State $v \gets \textsc{LLMVerify}(c,E,\mathrm{evidence\_only}=\mathrm{True})$
        \Comment{Check support from retrieved passages.}
    \If{$v = \mathrm{YES}$}
        \State $\mathrm{supported} \gets \mathrm{supported}+1$
    \EndIf
\EndFor
\State $\rho \gets \mathrm{supported}/|A|$
    \Comment{Fraction of claims supported by the source}
\State \Return $(\rho,\rho \geq \tau)$

\end{algorithmic}
\end{algorithm}

\textbf{Benchmark Reconstruction.}
We build SoundnessBench in five steps, shown in Fig.~\ref{fig:pipeline}.

\begin{itemize}[leftmargin=*,itemsep=0.1in, topsep=0in, parsep=0in]
    \item \textbf{Step 1: Data Collection.}
    We collect ICLR papers spanning 39 primary research areas such as RL, generative modeling, XAI, and applications (39 primary areas in total). We start from 2022 because earlier releases do not consistently provide reviewer soundness (or closely related) scores needed for filtering and label derivation. This pool provides coverage across years and research areas. We exclude desk-rejected papers, since their rejections may reflect factors other than scientific soundness. To reduce label noise, we retain papers with mean reviewer confidence of at least 3 out of 5 and a standard deviation of normalized soundness scores below 15\%.
    
    \item \textbf{Step 2: ICLR Proxy Label Assignment.}
    We assign labels from reviewer \emph{soundness} sub-scores rather than overall ratings or novelty/presentation scores. Papers with mean scores $\ge 3$ are labeled high soundness; those $\le 2$ are labeled low soundness. Fractional scores arise from averaging individual reviewer ratings. We exclude intermediate cases to improve class separation. These labels serve as proxies for proposal-stage methodological validity, since reviewers assessed full papers and may rely on results or other evidence absent from the extracted proposals.

    \item \textbf{Step 3: Proposal Extraction.}
    From the papers filtered in Step 2, we extract structured research proposals (abstract, related work, risks, hypothesis, and experimental design) directly from the source PDFs, following \citet{yamada2025aiscientistv2}. We use Gemini 2.5 Pro with a task-specific prompt that instructs the model to exclude experimental results, direct outcome claims, and acceptance cues, preserve near-verbatim wording, and avoid adding information unsupported by the source paper (Appx.~\ref{sec:supp:prompt-extract}). Example extractions are provided in Appx.~\ref{sec:supp:proposal-examples}.
    
    \item \textbf{Step 4: Extraction Verification.}
    To reduce extraction errors, we verify extracted claims against their source PDFs (Alg.~\ref{alg:verification}). Following prior work on atomic factuality and evidence-grounded verification~\citep{min-etal-2023-factscore,thorne2018fever}, we decompose each hypothesis-experiment pair into individual claims, retrieve relevant passages, and check whether they support each claim (Appx.~\ref{sec:supp:prompt-decompose}, Appx.~\ref{sec:supp:prompt-verify}, and Appx.~\ref{sec:supp:atomic-example}). We retain extractions with a supported-claim fraction of at least $\tau=0.7$, using chunk size $\ell=3000$, overlap $o=200$, and retrieval depth $k=3$. Overall, 66.93\% of candidates pass this filter; Fig.~\ref{fig:stats} reports the mean support ratio. This check assesses extraction fidelity rather than methodological soundness or label correctness.

    \item \textbf{Step 5: Final Benchmark Dataset.}
    The final benchmark contains 1,099 research proposals, including 458 low-soundness and 641 high-soundness instances according to the ICLR-derived proxy labels. Detailed benchmark statistics are presented in Fig.~\ref{fig:stats}.
\end{itemize}

\textbf{Proposal-Only Expert Annotation.}
To complement the ICLR-derived proxy labels, we recruit 12 machine learning researchers and assign each proposal to one expert annotator given their domain expertise in the relevant research areas. Annotators provide a sound/unsound judgment using only the hypothesis and proposed experimental plan (the same input given to models) without seeing the ICLR-derived label or experimental results. Given the time and domain expertise required, annotation cost and researcher availability limit coverage to 327 of the 1,099 proposals. We additionally identify the subset where expert judgments agree with the ICLR-derived labels for further analysis.

\section{Evaluation}

\subsection{Evaluation Setup}\label{sec:eval:setup}

\textbf{Models.}
We evaluate \textbf{12 frontier models} spanning small and large scales, non-reasoning and reasoning, and both closed- and open-source models. We include GPT-4o, GPT-5.4-Mini, GPT-5.4, Claude-Opus-4.6, Claude-Sonnet-4.6, Gemini-2.5-Pro, Gemini-3-Flash, Gemini-3.1-Pro, Qwen-3.5-27B, Qwen-3.5-122B-A10B, LLaMA-3.3-70B, Kimi-Linear-48B-A3B.

\textbf{Inference Setup.} We run closed-source models through provider APIs and serve open-source models with vLLM on two NVIDIA H200 GPUs. We use \texttt{max\_tokens=8192} and \texttt{temperature=0.2}; other settings follow provider or framework defaults.

\textbf{Evaluation Protocol.}
We instruct LLMs to classify SoundnessBench proposals as high or low soundness, providing a detailed justification before the final label. We evaluate three settings: (1) \textit{Standard prompt}, which requests a structured assessment without an explicit conservative prior; (2) \textit{Aggressive prompt}, which instructs models to default to low soundness unless the idea and experimental design are clearly strong and well-justified; and (3) \textit{Literature-grounded scientific agent}, using ScholarEval~\citep{moussa2025scholareval}, a retrieval-augmented evaluator. Data-construction prompts and additional evaluation analyses are in Appx.~\ref{sec:supp:data-construction} and Appx.~\ref{sec:supp:evaluation}, respectively.

\textbf{Metrics.} We report confusion matrices, class-wise recall (Low R and High R), and Macro F1 (the unweighted average of per-class F1 scores). Low R measures rejection of low-soundness proposals, while High R measures identification of high-soundness proposals. Taking high soundness as the positive class, false positives (low$\rightarrow$high) reflect optimism, whereas false negatives (high$\rightarrow$low) reflect over-conservatism relative to the reference labels.

\subsection{Main Results under ICLR Proxy Labels}\label{sec:eval:main-results}\label{Sec: Main Results under ICLR Proxy Labels}

\begin{figure}[ht]
    \centering
    \includegraphics[width=.88\linewidth]{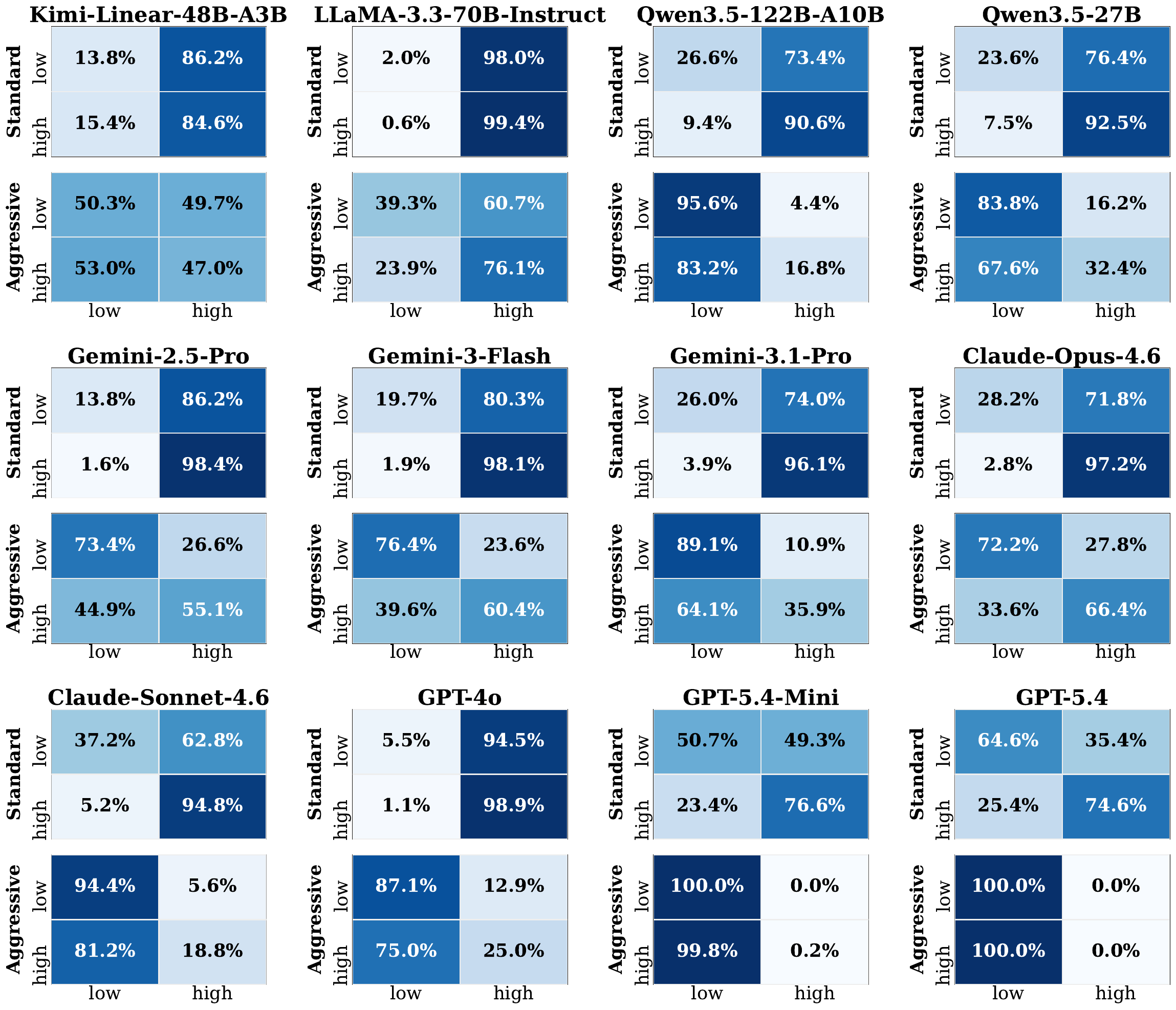}
    \caption{\textit{Soundness judgments under standard and aggressive prompting.} Each model panel shows row-normalized confusion matrices under standard (top) and aggressive (bottom) prompting on 1,099 proposals with ICLR-derived proxy labels. Rows indicate reference labels; columns indicate predictions. Aggressive prompting generally reduces false approvals while increasing false rejections.}\label{fig:eval-confusion-standard-agressive}
\end{figure}

We evaluate all 1,099 proposals using the ICLR-derived proxy labels, providing a large-scale assessment of model judgments. Tab.~\ref{tab:main_results} reports class-wise recall and Macro F1 under standard and aggressive prompting, and Fig.~\ref{fig:eval-confusion-standard-agressive} shows the corresponding confusion matrices.

\begin{table}[!htbp]
  \centering
  \caption{Summary evaluation metrics across 12 models under standard and aggressive
  prompting. Low R and High R denote recall on low- and
  high-soundness proposals. Macro F1 is the unweighted average of
  per-class F1 scores. Best Macro F1 per condition is \textbf{bolded}.
  $\dagger$ denotes reasoning model.}\label{tab:main_results}
  \resizebox{0.9\textwidth}{!}{%
  \begin{tabular}{lcccccc}
  \toprule
  & \multicolumn{3}{c}{\textbf{Standard Prompt}}
  & \multicolumn{3}{c}{\textbf{Aggressive Prompt}} \\
  \cmidrule(lr){2-4} \cmidrule(lr){5-7}
  \textbf{Model}
  & \textbf{Low R} $\uparrow$ & \textbf{High R} $\uparrow$ & \textbf{Macro F1} $\uparrow$
  & \textbf{Low R} $\uparrow$ & \textbf{High R} $\uparrow$ & \textbf{Macro F1} $\uparrow$ \\
  \midrule
  \multicolumn{7}{l}{\textit{Open-source Models}} \\
  LLaMA-3.3-70B                    & 2.0  & 99.4 & 38.8 & 39.3 & 76.1 & 57.4 \\
  Kimi-Linear-48B-A3B              & 13.8 & 84.6 & 44.5 & 50.3 & 47.0 & 48.1 \\
  Qwen3.5-27B$^\dagger$           & 23.6 & 92.5 & 55.0 & 83.8 & 32.4 & 52.7 \\
  Qwen3.5-122B-A10B$^\dagger$     & 26.6 & 90.6 & 56.4 & 95.6 & 16.8 & 44.7 \\
  \midrule
  \multicolumn{7}{l}{\textit{Closed-source Models}} \\
  GPT-4o$^\dagger$                 & 5.5  & 98.9 & 42.2 & 87.1 & 25.0 & 48.4 \\
  GPT-5.4-mini                     & 50.7 & 76.6 & 63.8 & 100.0 & 0.2 & 29.6 \\
  GPT-5.4$^\dagger$                & 64.6 & 74.6 & \textbf{69.6} & 100.0 & 0.0 & 29.4 \\
  Gemini-2.5-Pro$^\dagger$         & 13.8 & 98.4 & 49.7 & 73.4 & 55.1 & 62.7 \\
  Gemini-3-Flash                   & 19.7 & 98.1 & 54.5 & 76.4 & 60.4 & 67.0 \\
  Gemini-3.1-Pro$^\dagger$         & 26.0 & 96.1 & 58.4 & 89.1 & 35.9 & 56.9 \\
  Claude-Sonnet-4.6$^\dagger$      & 37.2 & 94.8 & 65.2 & 94.4 & 18.8 & 46.6 \\
  Claude-Opus-4.6$^\dagger$        & 28.2 & 97.2 & 60.2 & 72.2 & 66.4 & \textbf{69.1} \\
  \midrule
  \textbf{Mean}                    & 26.0 & 91.8 & 54.9 & 80.1 & 36.2 & 51.1 \\
  \bottomrule
  \end{tabular}%
  }
  \end{table}

\textbf{Across LLM families, we observe a pervasive optimism bias: models frequently rate proposals with low reviewer-derived soundness labels as sound.} Under standard prompting, mean low-soundness recall is 26.0\% (equivalently, a 74.0\% false-positive rate), while mean high-soundness recall remains 91.8\%. At the model level, 9/12 models (including Gemini-3.1-Pro and Claude-Opus-4.6) label more than 70\% of low-soundness proposals as high soundness. The strongest optimism appears in LLaMA-3.3-70B and GPT-4o, which classify 98.0\% and 94.5\% of low-soundness proposals as high soundness. This pattern indicates a broad tendency toward approval relative to the ICLR-derived proxy labels. Such permissive first-gate judgments could lead to unnecessary experimentation when approved proposals contain methodological flaws. Representative false-positive and true-positive examples, together with concrete model responses, are provided in Appx.~\ref{sec:supp:qualitative-fp-tp}.

\textit{A few frontier LLMs reduce optimism bias, but at the cost of missing more high-soundness proposals.} GPT-5.4 and GPT-5.4-mini achieve higher low-soundness recall (64.6\% and 50.7\%), alongside lower high-soundness recall (74.6\% and 76.6\%), than most other models. This suggests a rejection--identification trade-off in current frontier LLMs for proposal-stage soundness judgment. Stricter filtering can suppress false approvals, but it also increases the risk of rejecting promising ideas.

\textbf{Under Aggressive Prompting: Bias Shifts Toward Over-Conservatism.} We test how judgments change when the decision threshold is made deliberately stricter while keeping the same models and task: default to \textit{low soundness} unless the idea and experimental design are clearly strong and well-justified. As shown in Tab.~\ref{tab:main_results} and Fig.~\ref{fig:eval-confusion-standard-agressive}, models generally become more conservative. Mean low-soundness false positives drop from 74.0\% (standard prompt) to 19.9\% (aggressive prompt), and 10/12 models are below 30\%; meanwhile, mean high-soundness recall drops from 91.8\% to 36.2\%, with 7/12 models below 40\%. At the extreme, GPT-5.4 and GPT-5.4-Mini move to near-always-low behavior under aggressive prompting: low-soundness false positives are 0\% for both, while high-soundness recall falls to 0.0\% and 0.2\%, respectively. Other models show the same pattern in milder form. For example, Qwen3.5-122B-A10B and Claude-Sonnet-4.6 strongly reject low-soundness ideas (95.6\% and 94.4\% rejection), but correctly identify only 16.8\% and 18.8\% of high-soundness proposals. Consistent with this, mean Macro F1 decreases from 54.9 to 51.1.

\textit{Taken together, these results reveal prompt-sensitive error patterns
in proposal-stage soundness judgment.} Standard prompting favors approval, whereas aggressive prompting reduces false approvals at the cost of substantially more false rejections. The effect varies across models. GPT-5.4's Macro F1 falls from 69.6\% to 29.4\%, while Claude-Opus-4.6 improves from 60.2\% to 69.1\%. These results highlight the difficulty of balancing rejection and identification through prompt instructions alone.

\subsection{Evaluation against Expert Judgments}\label{sec:eval:expert}

A key concern is that ICLR-derived proxy labels may reflect experimental outcomes unavailable in the proposal-only task. We therefore evaluate the same 327 proposals using both their original ICLR-derived labels and proposal-only expert judgments, holding model predictions fixed to isolate the effect of label source. We additionally evaluate the subset where both label sources agree as a sensitivity check. In summary, our results show that the central error pattern persists across label sources: standard prompting favors approval, while aggressive prompting substantially increases false rejections.Detailed confusion matrices and results are in Appx.~\ref{sec:supp:expert-evaluation}, Appx.~\ref{sec:supp:iclr-matched-evaluation}, and Appx.~\ref{sec:supp:agreement-evaluation}.

\begin{table}[t]
  \centering
  \caption{\textbf{Evaluation across label sources.}
  ICLR and expert labels are evaluated on the same 327 annotated
  proposals with model predictions held fixed. The agreement subset
  contains proposals where both labels agree and serves as a sensitivity
  check. Low R and High R denote class-wise recall; all metrics are
  percentages. Mean rows average across models. Best Macro F1 within
  each group and prompting condition is \textbf{bolded}.
  $\dagger$ denotes reasoning model.}\label{tab:label-source-comparison}
  \setlength{\tabcolsep}{1pt}
  \renewcommand{\arraystretch}{1.08}
  \resizebox{.85\textwidth}{!}{%
  \begin{tabular}{lccccccccc}
  \toprule
  & \multicolumn{3}{c}{\textbf{ICLR Labels}}
  & \multicolumn{3}{c}{\textbf{Expert Labels}}
  & \multicolumn{3}{c}{\textbf{Agreement Subset}} \\
  \cmidrule(lr){2-4} \cmidrule(lr){5-7} \cmidrule(lr){8-10}
  \textbf{Model}
  & \textbf{Low R} $\uparrow$ & \textbf{High R} $\uparrow$ & \textbf{Macro F1} $\uparrow$
  & \textbf{Low R} $\uparrow$ & \textbf{High R} $\uparrow$ & \textbf{Macro F1} $\uparrow$
  & \textbf{Low R} $\uparrow$ & \textbf{High R} $\uparrow$ & \textbf{Macro F1} $\uparrow$ \\
  \midrule
  \multicolumn{10}{l}{\textbf{Standard Prompting}} \\
  \addlinespace[2pt]
  LLaMA-3.3-70B
  & 4.3 & 96.6 & 30.6 & 5.4 & 100.0 & 26.6 & 4.7 & 100.0 & 26.1 \\
  Kimi-Linear-48B-A3B
  & 17.3 & 84.0 & 39.3 & 16.2 & 81.4 & 32.7 & 17.3 & 81.2 & 33.7 \\
  Qwen3.5-27B$^\dagger$
  & 34.6 & 80.7 & 51.1 & 36.5 & 91.9 & 51.0 & 37.2 & 91.3 & 51.5 \\
  Qwen3.5-122B-A10B$^\dagger$
  & 38.0 & 77.3 & 52.2 & 41.1 & 91.9 & 54.3 & 40.3 & 92.8 & 54.1 \\
  GPT-4o$^\dagger$
  & 10.6 & 94.1 & 36.1 & 12.0 & 100.0 & 33.1 & 11.5 & 100.0 & 32.8 \\
  GPT-5.4-mini
  & 59.6 & 59.7 & 58.5 & 61.8 & 73.3 & 62.2 & 62.3 & 73.9 & 62.8 \\
  GPT-5.4$^\dagger$
  & 72.6 & 56.3 & 64.3 & 73.4 & 69.8 & \textbf{68.4}
  & 75.4 & 72.5 & \textbf{70.8} \\
  Gemini-2.5-Pro$^\dagger$
  & 24.5 & 92.4 & 47.5 & 24.9 & 100.0 & 44.3 & 26.7 & 100.0 & 45.9 \\
  Gemini-3-Flash
  & 29.8 & 92.4 & 51.6 & 28.6 & 97.7 & 46.7 & 31.9 & 98.6 & 49.6 \\
  Gemini-3.1-Pro$^\dagger$
  & 36.1 & 89.9 & 55.2 & 34.9 & 96.5 & 51.1 & 38.2 & 98.6 & 54.2 \\
  Claude-Sonnet-4.6$^\dagger$
  & 54.4 & 83.2 & \textbf{65.0} & 52.0 & 91.4 & 61.7 & 56.7 & 95.5 & 66.4 \\
  Claude-Opus-4.6$^\dagger$
  & 34.8 & 95.7 & 54.7 & 35.4 & 100.0 & 54.9 & 38.1 & 100.0 & 55.9 \\
  \midrule
  \rowcolor{gray!12}
  \textbf{Mean}
  & 34.7 & 83.5 & 50.5 & 35.2 & 91.1 & 48.9 & 36.7 & 92.0 & 50.3 \\
  \midrule
  \multicolumn{10}{l}{\textbf{Aggressive Prompting}} \\
  \addlinespace[2pt]
  LLaMA-3.3-70B
  & 52.9 & 61.3 & 55.4 & 55.6 & 74.4 & 58.7 & 56.0 & 72.5 & 58.4 \\
  Kimi-Linear-48B-A3B
  & 52.4 & 47.9 & 49.5 & 53.1 & 50.0 & 48.8 & 53.4 & 47.8 & 48.3 \\
  Qwen3.5-27B$^\dagger$
  & 89.9 & 20.2 & 52.8 & 92.1 & 30.2 & 62.3 & 92.1 & 29.0 & 61.5 \\
  Qwen3.5-122B-A10B$^\dagger$
  & 96.2 & 12.6 & 49.6 & 96.3 & 16.3 & 55.4 & 96.9 & 17.4 & 56.5 \\
  GPT-4o$^\dagger$
  & 89.4 & 18.5 & 51.4 & 89.6 & 22.1 & 55.8 & 90.1 & 23.2 & 56.7 \\
  GPT-5.4-mini
  & 100.0 & 0.0 & 38.9 & 100.0 & 0.0 & 42.4 & 100.0 & 0.0 & 42.4 \\
  GPT-5.4$^\dagger$
  & 100.0 & 0.0 & 38.9 & 100.0 & 0.0 & 42.4 & 100.0 & 0.0 & 42.4 \\
  Gemini-2.5-Pro$^\dagger$
  & 81.2 & 44.5 & \textbf{63.3} & 80.5 & 52.3 & 66.0 & 83.8 & 53.6 & 68.8 \\
  Gemini-3-Flash
  & 81.2 & 43.7 & 62.8 & 80.9 & 52.3 & 66.3 & 83.2 & 55.1 & \textbf{69.1} \\
  Gemini-3.1-Pro$^\dagger$
  & 92.3 & 26.1 & 58.0 & 91.7 & 31.4 & 62.7 & 92.7 & 36.2 & 66.1 \\
  Claude-Sonnet-4.6$^\dagger$
  & 94.0 & 4.0 & 42.3 & 96.2 & 9.1 & 48.8 & 95.5 & 6.2 & 46.8 \\
  Claude-Opus-4.6$^\dagger$
  & 81.2 & 41.4 & 61.7 & 83.6 & 52.0 & \textbf{67.8} & 83.9 & 55.6 & 68.9 \\
  \midrule
  \rowcolor{gray!12}
  \textbf{Mean}
  & 84.2 & 26.7 & 52.0 & 85.0 & 32.5 & 56.5 & 85.6 & 33.0 & 57.2 \\
  \bottomrule
  \end{tabular}%
  }
  \end{table}

\textbf{The optimism bias persists under proposal-only expert labels.} In Tab.~\ref{tab:label-source-comparison}, under standard prompting, mean low-soundness recall is nearly unchanged: 34.7\% using matched ICLR labels and 35.2\% using expert labels, corresponding to false-positive rates of 65.3\% and 64.8\%, respectively. Mean high-soundness recall increases from 83.5\% to 91.2\% under expert labels. On the agreement subset, low- and high-soundness recall are 36.7\% and 92.1\%. Thus, the tendency to approve low-soundness proposals persists when ICLR-derived labels are replaced with proposal-only expert judgments.

\textbf{Aggressive prompting again exchanges false approvals for false rejections.} Mean low-soundness recall rises to 84.3\%, 85.0\%, 85.6\% under matched ICLR labels, expert labels, their agreement subset, respectively. However, high-soundness recall falls to 26.7\%, 32.6\%, 33.0\%. Aggressive prompting improves mean Macro F1 under expert labels from 48.7\% to 56.6\%. These results indicates better overall classification performance alongside a substantial shift in the error balance. 

\subsection{Additional Analyses and Robustness Checks}\label{sec:eval:controls}

We provide additional analysis and robustness checks. Unless otherwise stated, these analyses use ICLR-derived proxy labels. In summary, our results show that the observed optimism bias is unlikely to be explained solely by missing literature context, model scale, instruction tuning, public corpus contamination, shallow structural features. These findings highlight subtle limitations in reliable proposal-stage soundness judgment by LLMs.

\begin{itemize}[leftmargin=*,itemsep=0.04in,topsep=0.03in,parsep=0in]
    \item \textbf{Literature grounding agent does not eliminate optimism.} ScholarEval~\citep{moussa2025scholareval}, a retrieval-augmented scientific-idea evaluator, predicts nearly every proposal as high soundness at thresholds 5.0 to 7.0. On the coverage $\geq 0.8$ subset, raising the threshold to 8.0 yields 34.3\% low-soundness recall, 73.3\% high-soundness recall, and 53.7\% Macro F1. This is more balanced and achieves higher Macro F1 than matched standard (48.8\%) and aggressive prompting (43.8\%), but still approves 65.7\% of low-soundness proposals (Appx.~\ref{sec:supp:scholareval}).

    \item \textbf{Neither scale nor instruction tuning consistently corrects the bias.} Within the Qwen3.5 family, scaling from 2B to 35B increases high-soundness recall from 71.8\% to 92.8\%, but decreases low-soundness recall from 31.0\% to 19.2\% (Appx.~\ref{sec:supp:scale-bias}). The Qwen3.5-35B base and instruction-tuned variants also behave similarly under the standard prompt (Appx.~\ref{sec:supp:instruction-tuning-bias}). These suggests that the optimism bias is not solely a small-model or instruction-tuning artifact.

    \item \textbf{Contamination and paper recognition are unlikely to explain the pattern.} On an ICLR 2026-only split evaluated with cutoff-restricted models, the false-positive rate remains comparable to the full dataset (77.47\% vs.\ 73.88\%) (Appx.~\ref{sec:supp:contamination-control}). Masking method names and other identifying phrases changes aggregate predictions by roughly one percentage point (Appx.~\ref{sec:supp:identifier-removal})

    \item \textbf{Shallow structural features fail differently from the evaluated LLMs.} We try simple threshold baselines based on proposal length, experiment count, and risk-factor count in Appx.~\ref{sec:supp:surface-baselines}. These baselines tend to over-reject high-soundness proposals, whereas LLMs over-approve low-soundness proposals. Both approaches show failures but in different ways. 

    \item \textbf{GPT-5.4 detects severe flaws but remain insensitive to subtler ones.} Injecting severe hypothesis--experiment mismatches into 100 high-soundness proposals reduces GPT-5.4's approval rate from 77.0\% to 1.0\%, showing that the model attends to salient methodological content (Appx.~\ref{sec:supp:adversarial-injection}). 
\end{itemize}

\section{Related Work}\label{sec:app:related-work}

\textbf{Autonomous AI Research Agents.}
Recent AI scientist agents aim to automate large parts of the research pipeline. \textit{The AI Scientist}~\citep{lu2024aiscientist} and \textit{The AI Scientist-v2}~\citep{yamada2025aiscientistv2} are early systems that attempt this process by autonomously generating ideas, writing code, running experiments, and drafting papers. \textit{Agent Laboratory}~\citep{schmidgall2025agentlab} structures this process into literature review, experimentation, and report writing, with optional human feedback. Beyond papers, \textit{autoresearch}~\citep{karpathy2026autoresearch} provides a simple open-source loop where an agent edits code, runs short real experiments, and keeps changes only when metrics improve. These systems show that LLM agents can execute multiple parts of research workflows. However, their evaluations do not directly test whether LLMs can judge whether a hypothesis and experimental design are methodologically sound \emph{before} execution begins. Like human researchers triaging projects, this first-gate decision determines whether time and compute should be invested.

\textbf{Evaluating Research Agents and Idea Quality.} \cite{si2024llmideas} show that LLM-generated ideas can be novel while being weaker on feasibility, and that the gap between proposal-stage promise and post-execution outcomes can widen after implementation~\citep{si2025ideationexecution}. We take this implication seriously. Proposal-only judgment may be intrinsically limited because some weaknesses or strengths become visible only after experiments are run. SoundnessBench is therefore not intended to predict final research impact or full-paper acceptance. Instead, it asks a narrower diagnostic question: how much reviewer-assigned methodological soundness is recoverable from proposal-stage evidence, and whether LLMs can act as stable first-gate critics under that limitation. At the system level, most current benchmarks focus on evaluating execution~\citep{chan2024mle,starace2025paperbench,wu2025innovatorbench,lupidi2026airs,garikaparthi2026researchgym}. Additionally, a few idea-evaluation benchmarks target dimensions such as impact and novelty~\citep{jiang2026hindsight, rinobench2026}, or model scientific taste from community-level signals such as publication venue and citation outcomes~\citep{tong2026scientifictaste}. These directions are important, but their targets differ from ours.

\textbf{LLMs in Peer-Review Workflows.}
Existing work studies LLMs in workflows for reviewing completed papers. LLM assistance can improve review clarity and actionability in human review workflows~\citep{thakkar2025llmfeedback}. Related work also studies LLM-enhanced reviewer recommendation~\citep{omnireview2026} and AI-text detection in peer reviews~\citep{isyourpaperreviewed2025}. However, this literature evaluates \emph{post hoc} review, where the paper, results, and framing are already available. SoundnessBench instead studies \emph{pre-execution} judgment from the research proposal alone.

\textbf{LLM Sycophancy and Fragile Judgment.}
Our work is connected to studies of LLM sycophancy and fragile judgment. Previous work shows that LLMs can favor answers matching a user's stated beliefs over truthful answers~\citep{sharma2024sycophancy}. Other work jointly examines sycophancy, skepticism, and miscalibrated refusal in causal reasoning~\citep{causalt5k}. Prompt-framing studies show that small syntactic changes can flip model decisions in ethical decision tasks even when content is unchanged~\citep{syntacticfragility}. These suggest that scientific criticism is not only a knowledge problem, but also a robustness problem. We test both.
\section{Conclusion}

We introduced SoundnessBench to evaluate a core capability for autonomous ML research agents: deciding whether a hypothesis and experimental design appear methodologically sound before running expensive experiments. Across 12 frontier LLMs, standard prompting produces a strong optimism bias, with a 74.0\% mean false-positive rate. Aggressive prompting lowers this rate to 19.9\% but reduces high-soundness recall to 36.2\%, shifting rather than resolving the error. This pattern persists under proposal-only expert labels and across controls for literature grounding, public-corpus contamination, surface features, and injected methodological flaws. At this early gate in the research pipeline, such instability can misallocate compute and attention. We view SoundnessBench as a step toward safer and more useful AI research agents, and as a tool for diagnosing where model calibration breaks down. The observed optimism tendency and prompt sensitivity suggest that reliable proposal-stage scientific judgment will likely require targeted training, calibration, or human-in-the-loop review beyond prompting alone.

\textbf{Limitations.} The large-scale labels are derived from reviewer soundness sub-scores and remain imperfect proxies because reviewers assessed complete papers, including their results and presentation. Proposal-only expert judgments on 327 proposals reduce this concern but do not establish a human performance ceiling. SoundnessBench is also limited to ICLR machine-learning research and should not be interpreted as measuring scientific judgment across all domains. Because the source corpus is public, contamination cannot be eliminated completely, and the scale and instruction-tuning analyses cover only one model family. Future work should expand expert annotation, evaluate private or continuously refreshed proposals, cover additional scientific fields and model families, and connect proposal-stage judgments with downstream execution outcomes.





\section*{Acknowledgments}

Ho, Liu, and Huang are supported by DARPA HR001124S0029-AIQ-FP-019 and the National Science Foundation TRAILS Institute (2229885). Private support was provided by Open Philanthropy and Apple. The authors acknowledge the National Artificial Intelligence Research Resource (NAIRR) Pilot for supporting this research.

\bibliographystyle{plainnat}
\bibliography{references}

\appendix

\onecolumn
\section*{Appendix}

\section{Supplementary Material for Data Construction}
\label{sec:supp:data-construction}

\subsection{Prompt to Extract Proposal}\label{sec:supp:prompt-extract}\label{Sec: Prompt to Extract Proposal}

The prompt is designed to extract proposal-level content from each paper while explicitly excluding results and conclusions. The proposal format follows~\cite{yamada2025aiscientistv2}.

\begin{tcolorbox}[title=\textbf{System Prompt to Extract Proposal}]
\small

You are an expert at extracting structured research information from machine learning papers. Extract the research proposal in \textit{AI Scientist v2} format, capturing the intended hypothesis and experimental design \textbf{before any experimental results}. The goal is to represent the research plan and claims prior to execution, not confirmed findings.

\textbf{Output:} Return \textbf{valid JSON only}. Do not include markdown or explanations.

\vspace{0.3em}
\textbf{Global Principles}
\begin{itemize}
\item Extract only proposal-level content; exclude results and conclusions.
\item Prefer verbatim spans from the paper when possible.
\item Do not hallucinate or introduce structure not present in the paper.
\item If uncertain, omit the information.
\item Keep claims atomic and directly grounded in the source text.
\end{itemize}

\vspace{0.3em}
\textbf{Hypothesis}
\begin{itemize}
\item Extract from Introduction or Method sections.
\item Describe what the paper proposes or aims to test (1--2 sentences).
\item Prefer verbatim phrasing.
\end{itemize}

\textbf{Forbidden:} ``we show'', ``we demonstrate'', ``we find'', ``results show'', ``outperforms'' \\
\textbf{Allowed:} ``we propose'', ``we hypothesize'', ``we argue'', ``our approach aims to''

\vspace{0.3em}
\textbf{Experiments}
\begin{itemize}
\item Describe experimental plan only (no results or numbers).
\item Each entry must correspond to a distinct experiment group.
\end{itemize}

\textbf{Description:} goal of the experiment. \\
\textbf{Method:} datasets, models, baselines, setup (preserve exact names). \\
\textbf{Metrics:} list evaluation metrics (no values).

\vspace{0.3em}
\textbf{Related Work}
\begin{itemize}
\item Summarize prior work and research gap.
\item Do not include improvement claims.
\end{itemize}

\vspace{0.3em}
\textbf{Risk Factors}
\begin{itemize}
\item Extract explicit limitations from the paper.
\item Do not infer or invent new risks.
\end{itemize}

\vspace{0.3em}
\textbf{Abstract}
\begin{itemize}
\item Keep only problem, motivation, and proposed method.
\item Remove results, numbers, and comparisons.
\end{itemize}

\end{tcolorbox}

\begin{tcolorbox}[title=\textbf{User Prompt to Extract Proposal}]
\small

\textbf{Input}
\begin{itemize}
\item \textbf{Abstract:}
\begin{verbatim}
{abstract}
\end{verbatim}
\item The attached file is the \textbf{full paper PDF}.
\end{itemize}

Use both the abstract and the PDF to extract the research idea.

\vspace{0.3em}
\textbf{Instructions}
\begin{itemize}
\item Extract the \textbf{Short Hypothesis} from the \textbf{Introduction or Method sections}, not from results or conclusions.
\item All \textbf{Experiment fields} must reflect the \textbf{experimental plan}, not observed outcomes.
\item Preserve \textbf{dataset names, tasks, and baselines} when describing experiments.
\item Create \textbf{multiple experiment entries} if the paper evaluates multiple tasks or benchmarks.
\end{itemize}

\vspace{0.3em}
\textbf{Output Format (JSON)}
\begin{verbatim}
{
  "Name": "<short slug identifying the research contribution>",
  "Title": "<exact paper title>",
  "Short Hypothesis": "<proposed claim or method---no outcomes>",
  "Related Work": "<prior work and research gap>",
  "Abstract": "<problem + motivation + proposed method only>",
  "Experiments": [
    {
      "Description": "<goal of this experiment group>",
      "Method": "<datasets/tasks, model setup, baselines>",
      "Evaluation Metrics": "<metrics used>"
    }
  ],
  "Risk Factors and Limitations": [
    "<explicit limitation sentence from the paper>"
  ]
}
\end{verbatim}

\vspace{0.3em}
\textbf{Critical Constraints}
\begin{itemize}
\item Do \textbf{not} include results or conclusions.
\item Do \textbf{not} include numerical values or measured performance.
\item Do \textbf{not} reveal whether the hypothesis was confirmed.
\item Do \textbf{not} generalize dataset names.
\item Do \textbf{not} merge distinct experiments into a single entry.
\end{itemize}

\end{tcolorbox}

\subsection{Prompt to Decompose Atomic Claims from Proposal}\label{sec:supp:prompt-decompose}\label{Sec: Prompt to Decompose Atomic Claims from Proposal}

This prompt decomposes each proposal into minimal, independently verifiable factual claims so that each statement can be assessed clearly and grounded in source text. The detailed process is presented in Alg.~\ref{alg:verification}. We compute extraction faithfulness with a retrieval-backed atomic-claim score by decomposing each hypothesis\textendash{}experiment pair into atomic claims, retrieving supporting passages, and verifying each claim against the source paper, following prior work on atomic factuality and evidence-grounded verification~\citep{min-etal-2023-factscore,thorne2018fever}. We use GPT-5.4 for this process.

\begin{tcolorbox}[title=\textbf{System Prompt to Decompose Atomic Claims from Proposal}]
\small

You are a precise scientific analyst. Your task is to decompose a research hypothesis and experimental plan into atomic, independently verifiable claims.

\end{tcolorbox}

\vspace{1.0cm}

\begin{tcolorbox}[title=\textbf{User Prompt to Decompose Atomic Claims from Proposal}]
\small

\textbf{Task.} Decompose the following extraction into \textbf{atomic factual claims} that are directly verifiable from the paper text.

\vspace{0.3em}
\textbf{Rules}
\begin{itemize}
\item Output \textbf{one complete claim per line}.
\item Do not include numbering, bullets, prefixes, section headers, or commentary.
\item Do not output incomplete fragments.
\item Do not introduce synthetic labels (e.g., ``Experiment 1'', ``Experiment 2'').
\item Keep claims literal and close to wording likely present in the paper.
\end{itemize}

\vspace{0.3em}
\textbf{Scope of Claims}
Include only claims about:
\begin{itemize}
\item proposed method or hypothesis
\item planned evaluation setup
\item datasets, tasks, and baselines
\item evaluation metrics
\end{itemize}

Exclude:
\begin{itemize}
\item any result or outcome claims (e.g., ``improves'', ``outperforms'', ``achieves state-of-the-art'')
\end{itemize}

\vspace{0.3em}
\textbf{Input Extraction}
\begin{verbatim}
{extraction_proposal}
\end{verbatim}

\end{tcolorbox}

\subsection{Prompt to Verify Atomic Claims}\label{sec:supp:prompt-verify}

This prompt evaluates each claim\textendash{}evidence pair and produces a concise YES/NO judgment so that verification decisions remain explicit, consistent, and grounded in the provided textual evidence. We use Gemini-2.5-Flash for this process.

\vspace{1.0cm}

\begin{tcolorbox}[title=\textbf{System Prompt to Verify Atomic Claims}]
\small

You are a rigorous scientific fact-checker. Your task is to verify whether a claim is \textbf{directly supported} by the provided evidence from a paper.

\end{tcolorbox}

\begin{tcolorbox}[title=\textbf{User Prompt for Atomic Claim Verification}]
\small

\textbf{Input Claim}
\begin{verbatim}
{claim}
\end{verbatim}

\textbf{Evidence (excerpts from the paper)}
\begin{verbatim}
{evidence}
\end{verbatim}

\vspace{0.3em}
\textbf{Task}

Determine whether the claim is \textbf{directly supported} by the provided evidence.

\vspace{0.3em}
\textbf{Output Format}

Reply with exactly:
\begin{verbatim}
YES/NO: <one-sentence explanation>
\end{verbatim}

\end{tcolorbox}

\subsection{Example of Extracted Proposals}\label{sec:supp:proposal-examples}

We provide two extracted-proposal examples from SoundnessBench: one high-soundness case and one low-soundness case. The benchmark contains 458 low-soundness and 641 high-soundness instances in total. The format of these proposals follows \citep{yamada2025aiscientistv2}.

\begin{tcolorbox}[title=\textbf{Example: High-Soundness Proposal}]
\small
\raggedright{}

\textbf{Paper id}: FUaDMRVrbS \\

\textbf{Identifiability for Gaussian Processes with Holomorphic Kernels} \\

\textbf{Year}: 2025 \\

\textbf{Short Hypothesis} \\
A novel theoretical framework, based on the property of kernels being holomorphic around zero, can determine the identifiability of parameters in a wide range of GP kernels (e.g., squared exponential, periodic, rational quadratic) and their complex combinations, filling a gap left by methods that only apply to Matérn-type kernels. \\

\textbf{Related Work} \\
\textit{[Content omitted for brevity]} \\

\textbf{Abstract} \\
\textit{[Content omitted for brevity]} \\

\textbf{Experiments}

\textit{Experiment 1} \\
\textbf{Description:} To empirically support the theoretical results on parameter identifiability for several individual holomorphic kernels. \\
\textbf{Method:} The experiment involves estimating parameters for individual Squared Exponential (SE), Damped Periodic (DPer), Periodic (Per), Rational Quadratic (RQ), and Cosine kernels. This is done by fitting a GP model to synthetically generated data of increasing sample sizes ($n \in \{500, 1000, 2000, 5000\}$) with added Gaussian noise. The parameters are estimated using Maximum Likelihood Estimators (MLEs). \\
\textbf{Evaluation Metrics:} Convergence of the Maximum Likelihood Estimators (MLEs) for each parameter over 100 replicates.

\vspace{0.5em}
\textit{Experiment 2} \\
\textbf{Description:} To empirically support the theoretical results on parameter identifiability for a complex combined kernel, specifically the one from Equation (2) motivated by the Mauna Loa CO2 dataset. \\
\textbf{Method:} A GP with the combined kernel from Equation (2) is fitted to synthetically generated data designed to mimic the properties of the Mauna Loa dataset over a 45-year time interval. The experiment uses increasing sample sizes ($n \in \{50, 100, 200, 500\}$), and the Maximum Likelihood Estimators (MLEs) for the kernel's 10 parameters are computed. \\
\textbf{Evaluation Metrics:} Convergence of the Maximum Likelihood Estimators (MLEs) for each parameter in the combined kernel over 100 replicates. \\

\textbf{Risk Factors and Limitations}
\begin{itemize}
\item First, while establishing the identifiability of kernel parameters is a critical step, it does not necessarily guarantee the consistency of the MLE\@.
\item Second, extending our theoretical framework to encompass non-stationary kernels \ldots{} remains a largely open problem.
\item Third, another intriguing direction involves extending findings to infinitely differentiable kernels not holomorphic near 0.
\end{itemize}

\vspace{0.3em}
\textbf{Scores:}
\begin{itemize}
  \item Soundness: 3.25
  \item Support Ratio: 1.0
  \item Rigor Bucket: high
\end{itemize}

\end{tcolorbox}

\begin{tcolorbox}[title=\textbf{Example: Low-Soundness Proposal}]
\small
\raggedright{}

\textbf{Paper id}: h9ThYkkgSD4 \\

\textbf{Activation Function: Absolute Function,One Function Behaves more Individualized} \\

\textbf{Year:} 2023 \\

\textbf{Short Hypothesis} \\
Using the absolute value function, $y = |x|$, as a neural network activation will result in more ``individualized'' representations by preserving information from negative inputs, unlike ReLU\@. This property is hypothesized to make it more suitable for generation tasks and potentially less prone to overfitting compared to ReLU and Leaky ReLU.\\

\textbf{Related Work} \\
\textit{[Content omitted for brevity]} \\

\textbf{Abstract} \\
\textit{[Content omitted for brevity]} \\

\textbf{Experiments}

\textit{Experiment 1} \\
\textbf{Description:} To evaluate the performance of the absolute activation function against standard activations in a fully-connected neural network on an image classification task.\\
\textbf{Method:} A 5-layer fully-connected neural network will be trained on the MNIST dataset. The performance of the absolute function will be compared against ReLU and Leaky ReLU as baselines. The experiment will use the Adam optimizer and SparseCategoricalCrossentropy loss. The effect of different batch sizes will also be investigated. \\
\textbf{Evaluation Metrics:} ['Training Accuracy', 'Validation Accuracy', 'Training Loss', 'Validation Loss'] 

\vspace{0.5em}
\textit{Experiment 2} \\
\textbf{Description:} To evaluate the performance of the absolute activation function in a convolutional neural network (CNN) on an image classification task and to visualize the resulting feature maps. \\
\textbf{Method:} A CNN with 3 convolutional layers and 2 fully-connected layers will be trained on the MNIST dataset. The absolute function will be compared against ReLU and Leaky ReLU baselines. The experiment will explore the effect of varying batch sizes (32, 64, 128). Additionally, the intermediate activations from the convolutional layers will be visualized for both the absolute function and ReLU to qualitatively compare the learned features. \\
\textbf{Evaluation Metrics:} ['Training Accuracy', 'Validation Accuracy', 'Training Loss', 'Validation Loss', 'Qualitative visualization of intermediate activations'] 

\vspace{0.5em}
\textit{Experiment 3} \\
\textbf{Description:} To test the hypothesis that the ``individualization'' property of the absolute function makes it well-suited for image generation tasks using an autoencoder. \\
\textbf{Method:} A functionally separate autoencoder, with a convolutional encoder and a corresponding decoder, will be trained on the MNIST dataset. The experiment will compare the image generation quality when using the absolute function in the encoder versus using a Leaky ReLU baseline. The model will use the Adam optimizer and MSE loss. \\
\textbf{Evaluation Metrics:} ['Qualitative comparison of generated output images'] \\

\textbf{Risk Factors and Limitations}
\begin{itemize}
\item In order to generalization, the individualization is the reason of shake, the accuracy may be good in some set and may be worse in some set.
\end{itemize}

\textbf{Scores:}
\begin{itemize}
  \item Soundness: 1.0
  \item Support Ratio: 1.0
  \item Rigor Bucket: low
\end{itemize}

\end{tcolorbox}

\subsection{Example of Extracted Atomic Claims and Verification}\label{sec:supp:atomic-example}

\begin{tcolorbox}[title=\textbf{Example: Atomic Claims from Extracted Proposal}]
\small

\textbf{Source:} Identifiability for Gaussian Processes with Holomorphic Kernels

\vspace{0.3em}
\textbf{Atomic Claims (excerpt)}

\begin{itemize}
\item A novel theoretical framework is based on kernels being holomorphic around zero.
\item The framework determines identifiability of parameters in a wide class of Gaussian Process kernels.
\item The framework applies to squared exponential, periodic, and rational quadratic kernels.
\item The framework applies to complex combinations of kernels.
\item The framework addresses a gap left by methods that only apply to Matérn-type kernels.
\item A Gaussian Process model is fitted to synthetically generated data with added Gaussian noise.
\item Parameters are estimated using Maximum Likelihood Estimators.
\item Experiments evaluate convergence of estimators as sample size increases.
\end{itemize}

\vspace{0.3em}
\textbf{Summary:} 17 atomic claims extracted; all verified as supported (support ratio = 1.0).

\end{tcolorbox}
\section{Supplementary Material for Evaluation}
\label{sec:supp:evaluation}

\subsection{Evaluation Prompt Templates}
\label{sec:supp:evaluation-prompts}

We provide the evaluation prompt templates for transparency. The standard prompt asks models to produce a structured scientific rationale before the final classification, so the main setting already includes an explicit deliberation-style instruction. The aggressive prompt keeps the same task but changes the default decision rule to require stronger evidence before assigning high soundness.

\begin{tcolorbox}[title=\textbf{System Prompt for Standard Evaluation}]
\small

You are an expert ML researcher and peer reviewer. Classify the scientific rigor bucket of a research idea and your assessment confidence from 1 to 5 from its hypothesis and experiment description.

Output the assessment as a JSON object, including a detailed step-by-step justification for the rigor bucket selected.

\end{tcolorbox}

\begin{tcolorbox}[title=\textbf{User Prompt for Standard Evaluation}]
\small

Classify this hypothesis-experiment pair into one rigor bucket:
\begin{itemize}[leftmargin=*,itemsep=0pt,topsep=0pt]
    \item \texttt{low}: Weak scientific contribution. Hypothesis is vague or trivial, experiments lack controls or baselines, metrics are weak, or methodology has fundamental flaws.
    \item \texttt{high}: Strong scientific contribution. Hypothesis is clear and meaningful. Experiments are rigorous, controlled, include appropriate baselines/ablations, and use suitable metrics.
\end{itemize}

Confidence Score Scale:
\begin{itemize}[leftmargin=*,itemsep=0pt,topsep=0pt]
    \item 1: You are unable to assess this paper and have alerted the ACs to seek an opinion from different reviewers.
    \item 2: You are willing to defend your assessment, but it is quite likely that you did not understand the central parts of the submission or that you are unfamiliar with some pieces of related work. Math/other details were not carefully checked.
    \item 3: You are fairly confident in your assessment. It is possible that you did not understand some parts of the submission or that you are unfamiliar with some pieces of related work. Math/other details were not carefully checked.
    \item 4: You are confident in your assessment, but not absolutely certain. It is unlikely, but not impossible, that you did not understand some parts of the submission or that you are unfamiliar with some pieces of related work.
    \item 5: You are absolutely certain about your assessment. You are very familiar with the related work and checked the math/other details carefully.
\end{itemize}

\textbf{HYPOTHESIS:} \texttt{\{hypothesis\}}

\textbf{EXPERIMENT:} \texttt{\{experiment\}}

Output format:
\begin{verbatim}
{
  "justification": "<Detailed step-by-step justification>",
  "rigor_bucket": <"low" or "high">,
  "confidence": <1-5 integer>
}
\end{verbatim}

Constraints:
\begin{itemize}[leftmargin=*,itemsep=0pt,topsep=0pt]
    \item \texttt{rigor\_bucket} must be a choice in [\texttt{\symbol{34}low\symbol{34}}, \texttt{\symbol{34}high\symbol{34}}]
    \item \texttt{confidence} must be an integer in [1, 5]
\end{itemize}
\end{tcolorbox}

\begin{tcolorbox}[title=\textbf{System Prompt for Aggressive Evaluation}]
\small

You are a strict ML area chair applying an aggressive rigor filter. Classify scientific rigor from a hypothesis and experiment description.

Default to ``low'' unless the evidence clearly demonstrates strong scientific rigor with concrete controls, strong baselines, appropriate metrics, and a credible evaluation plan.

Output valid JSON only with a detailed step-by-step justification.

\end{tcolorbox}

\begin{tcolorbox}[title=\textbf{User Prompt for Aggressive Evaluation}]
\small

Classify this hypothesis-experiment pair into one rigor bucket under an aggressive standard:
\begin{itemize}[leftmargin=*,itemsep=0pt,topsep=0pt]
    \item \texttt{low}: choose this unless there is clear, concrete, and compelling evidence of rigorous methodology.
    \item \texttt{high}: only if the plan is explicitly strong on hypothesis clarity, experimental controls, baselines/ablations, metric validity, and methodological credibility.
\end{itemize}

Aggressive policy:
\begin{itemize}[leftmargin=*,itemsep=0pt,topsep=0pt]
    \item Penalize missing controls, vague methods, missing or weak baselines, underspecified metrics, unclear evaluation protocol, or hand-wavy claims.
    \item If information is incomplete or ambiguous, prefer \texttt{low}.
    \item Use \texttt{high} only when justification is unambiguous.
\end{itemize}

Confidence Score Scale:
\begin{itemize}[leftmargin=*,itemsep=0pt,topsep=0pt]
    \item 1: You are unable to assess this paper and have alerted the ACs to seek an opinion from different reviewers.
    \item 2: You are willing to defend your assessment, but it is quite likely that you did not understand the central parts of the submission or that you are unfamiliar with some pieces of related work. Math or other details were not carefully checked.
    \item 3: You are fairly confident in your assessment. It is possible that you did not understand some parts of the submission or that you are unfamiliar with some pieces of related work. Math or other details were not carefully checked.
    \item 4: You are confident in your assessment, but not absolutely certain. It is unlikely, but not impossible, that you did not understand some parts of the submission or that you are unfamiliar with some pieces of related work.
    \item 5: You are absolutely certain about your assessment. You are very familiar with the related work and checked the math or other details carefully.
\end{itemize}

\textbf{HYPOTHESIS:} \texttt{\{hypothesis\}}

\textbf{EXPERIMENT:} \texttt{\{experiment\}}

Output format:
\begin{verbatim}
{
  "justification": "<Detailed step-by-step justification>",
  "rigor_bucket": <"low" or "high">,
  "confidence": <1-5 integer>
}
\end{verbatim}

Constraints:
\begin{itemize}[leftmargin=*,itemsep=0pt,topsep=0pt]
    \item \texttt{rigor\_bucket} must be a choice in [\texttt{\symbol{34}low\symbol{34}}, \texttt{\symbol{34}high\symbol{34}}]
    \item \texttt{confidence} must be an integer in [1, 5]
\end{itemize}
\end{tcolorbox}

\subsection{Evaluation against Expert Judgments}
\label{sec:supp:expert-evaluation}

This section will report the same per-model metrics and confusion-matrix analysis as Tab.~\ref{tab:main_results} and Fig.~\ref{fig:eval-confusion-standard-agressive}, using proposal-only expert judgments on the annotated subset as reference labels.

\begin{table}[ht]
    \centering
    \caption{Evaluation metrics against proposal-only expert judgments. Low R and High R denote recall on low- and high-soundness
    proposals. Macro F1 is the unweighted average of per-class F1 scores. Best Macro F1 per condition is \textbf{bolded}. $\dagger$ denotes
    reasoning model.}
    \label{tab:expert_results}
    \resizebox{0.9\textwidth}{!}{%
    \begin{tabular}{lcccccc}
    \toprule
    & \multicolumn{3}{c}{\textbf{Standard Prompt}}
    & \multicolumn{3}{c}{\textbf{Aggressive Prompt}} \\
    \cmidrule(lr){2-4} \cmidrule(lr){5-7}
    \textbf{Model}
    & \textbf{Low R} $\uparrow$ & \textbf{High R} $\uparrow$ & \textbf{Macro F1} $\uparrow$
    & \textbf{Low R} $\uparrow$ & \textbf{High R} $\uparrow$ & \textbf{Macro F1} $\uparrow$ \\
    \midrule
    \multicolumn{7}{l}{\textit{Open-source Models}} \\
    LLaMA-3.3-70B                & 5.4  & 100.0 & 26.6 & 55.6 & 74.4 & 58.7 \\
    Kimi-Linear-48B-A3B          & 16.2 & 81.4  & 32.7 & 53.1 & 50.0 & 48.8 \\
    Qwen3.5-27B$^\dagger$       & 36.5 & 91.9  & 51.0 & 92.1 & 30.2 & 62.3 \\
    Qwen3.5-122B-A10B$^\dagger$ & 41.1 & 91.9  & 54.3 & 96.3 & 16.3 & 55.4 \\
    \midrule
    \multicolumn{7}{l}{\textit{Closed-source Models}} \\
    GPT-4o$^\dagger$             & 12.0 & 100.0 & 33.1 & 89.6 & 22.1 & 55.8 \\
    GPT-5.4-mini                 & 61.8 & 73.3  & 62.2 & 100.0 & 0.0 & 42.4 \\
    GPT-5.4$^\dagger$            & 73.4 & 69.8  & \textbf{68.4} & 100.0 & 0.0 & 42.4 \\
    Gemini-2.5-Pro$^\dagger$     & 24.9 & 100.0 & 44.3 & 80.5 & 52.3 & 66.0 \\
    Gemini-3-Flash               & 28.6 & 97.7  & 46.7 & 80.9 & 52.3 & 66.3 \\
    Gemini-3.1-Pro$^\dagger$     & 34.9 & 96.5  & 51.1 & 91.7 & 31.4 & 62.7 \\
    Claude-Sonnet-4.6$^\dagger$  & 52.0 & 91.4  & 61.7 & 96.2 & 9.1 & 48.8 \\
    Claude-Opus-4.6$^\dagger$    & 35.4 & 100.0 & 54.9 & 83.6 & 52.0 & \textbf{67.8} \\
    \midrule
    \textbf{Mean}                & 35.2 & 91.1 & 48.9 & 85.0 & 32.5 & 56.5 \\
    \bottomrule
    \end{tabular}%
    }
  \end{table}

\begin{figure}[ht]
    \centering
    \includegraphics[width=.88\linewidth]{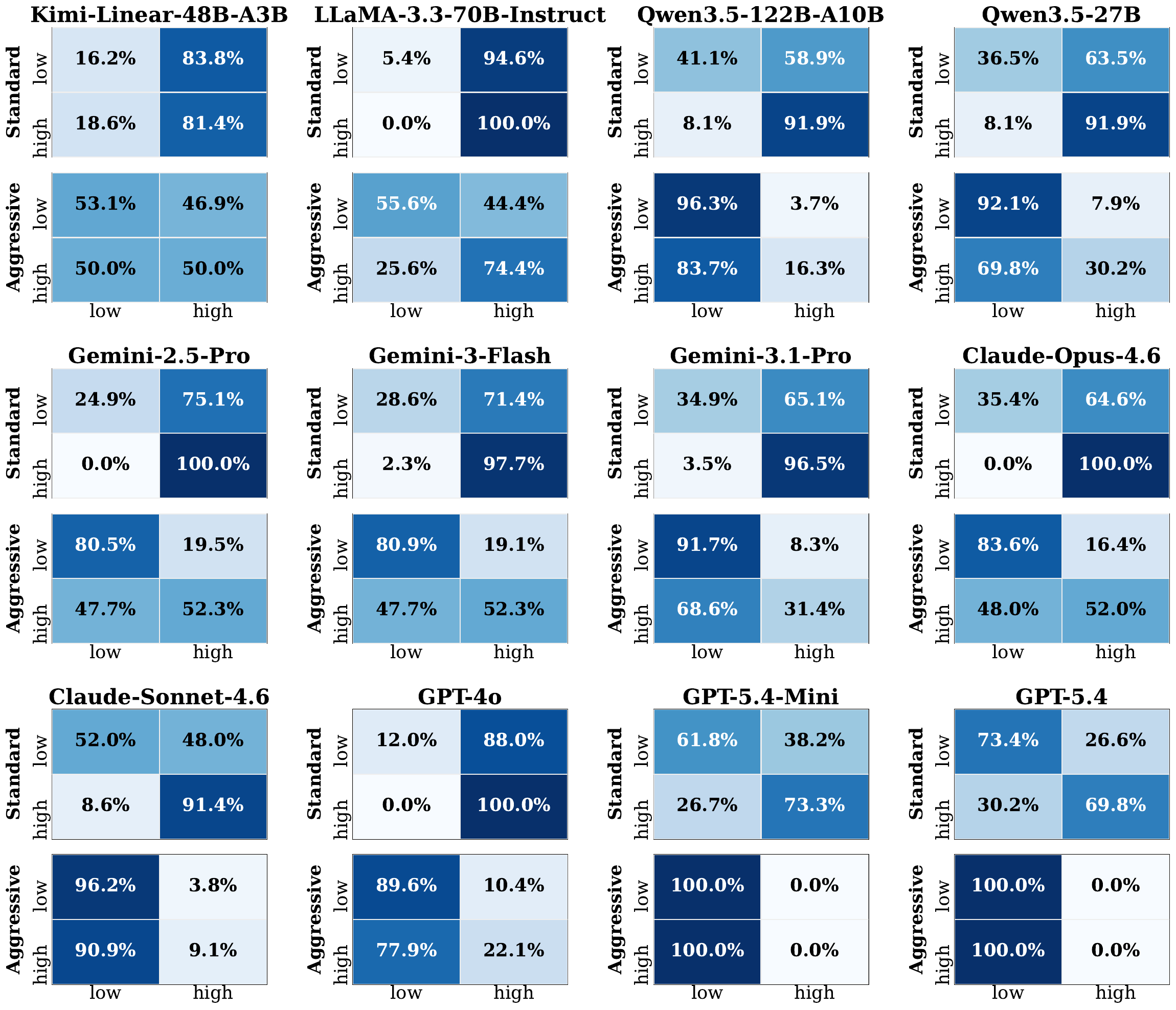}
    \caption{\textit{Soundness judgments against proposal-only expert labels under standard and aggressive prompting.} Each model panel shows row-normalized confusion matrices under standard (top) and aggressive (bottom) prompting on 327 expert-annotated proposals. Rows indicate expert reference labels; columns indicate predictions. Aggressive prompting generally reduces false approvals while increasing false rejections.}
    \label{fig:expert-confusion}
\end{figure}

\subsection{Evaluation on the Matched ICLR-Labeled Subset}
\label{sec:supp:iclr-matched-evaluation}

We evaluate the same proposals retained in the expert-relabeling setup using their original ICLR-derived proxy labels. Holding the proposals and model predictions fixed enables a direct comparison between evaluation under ICLR-derived and proposal-only expert labels.

\begin{table}[ht]
    \centering
    \caption{Evaluation metrics on the ICLR-labeled subset matched exactly to the proposals retained in the expert-relabeling setup. Low R
    and High R denote recall on low- and high-soundness proposals. Macro F1 is the unweighted average of per-class F1 scores. Best Macro F1
    per condition is \textbf{bolded}. $\dagger$ denotes reasoning model.}
    \label{tab:iclr_matched_results}
    \resizebox{0.9\textwidth}{!}{%
    \begin{tabular}{lcccccc}
    \toprule
    & \multicolumn{3}{c}{\textbf{Standard Prompt}}
    & \multicolumn{3}{c}{\textbf{Aggressive Prompt}} \\
    \cmidrule(lr){2-4} \cmidrule(lr){5-7}
    \textbf{Model}
    & \textbf{Low R} $\uparrow$ & \textbf{High R} $\uparrow$ & \textbf{Macro F1} $\uparrow$
    & \textbf{Low R} $\uparrow$ & \textbf{High R} $\uparrow$ & \textbf{Macro F1} $\uparrow$ \\
    \midrule
    \multicolumn{7}{l}{\textit{Open-source Models}} \\
    LLaMA-3.3-70B                & 4.3  & 96.6 & 30.6 & 52.9 & 61.3 & 55.4 \\
    Kimi-Linear-48B-A3B          & 17.3 & 84.0 & 39.3 & 52.4 & 47.9 & 49.5 \\
    Qwen3.5-27B$^\dagger$       & 34.6 & 80.7 & 51.1 & 89.9 & 20.2 & 52.8 \\
    Qwen3.5-122B-A10B$^\dagger$ & 38.0 & 77.3 & 52.2 & 96.2 & 12.6 & 49.6 \\
    \midrule
    \multicolumn{7}{l}{\textit{Closed-source Models}} \\
    GPT-4o$^\dagger$             & 10.6 & 94.1 & 36.1 & 89.4 & 18.5 & 51.4 \\
    GPT-5.4-mini                 & 59.6 & 59.7 & 58.5 & 100.0 & 0.0 & 38.9 \\
    GPT-5.4$^\dagger$            & 72.6 & 56.3 & 64.3 & 100.0 & 0.0 & 38.9 \\
    Gemini-2.5-Pro$^\dagger$     & 24.5 & 92.4 & 47.5 & 81.2 & 44.5 & \textbf{63.3} \\
    Gemini-3-Flash               & 29.8 & 92.4 & 51.6 & 81.2 & 43.7 & 62.8 \\
    Gemini-3.1-Pro$^\dagger$     & 36.1 & 89.9 & 55.2 & 92.3 & 26.1 & 58.0 \\
    Claude-Sonnet-4.6$^\dagger$  & 54.4 & 83.2 & \textbf{65.0} & 94.0 & 4.0 & 42.3 \\
    Claude-Opus-4.6$^\dagger$    & 34.8 & 95.7 & 54.7 & 81.2 & 41.4 & 61.7 \\
    \midrule
    \textbf{Mean}                & 34.7 & 83.5 & 50.5 & 84.2 & 26.7 & 52.0 \\
    \bottomrule
    \end{tabular}%
    }
  \end{table}

\begin{figure}[ht]
    \centering
    \includegraphics[width=.8\linewidth]{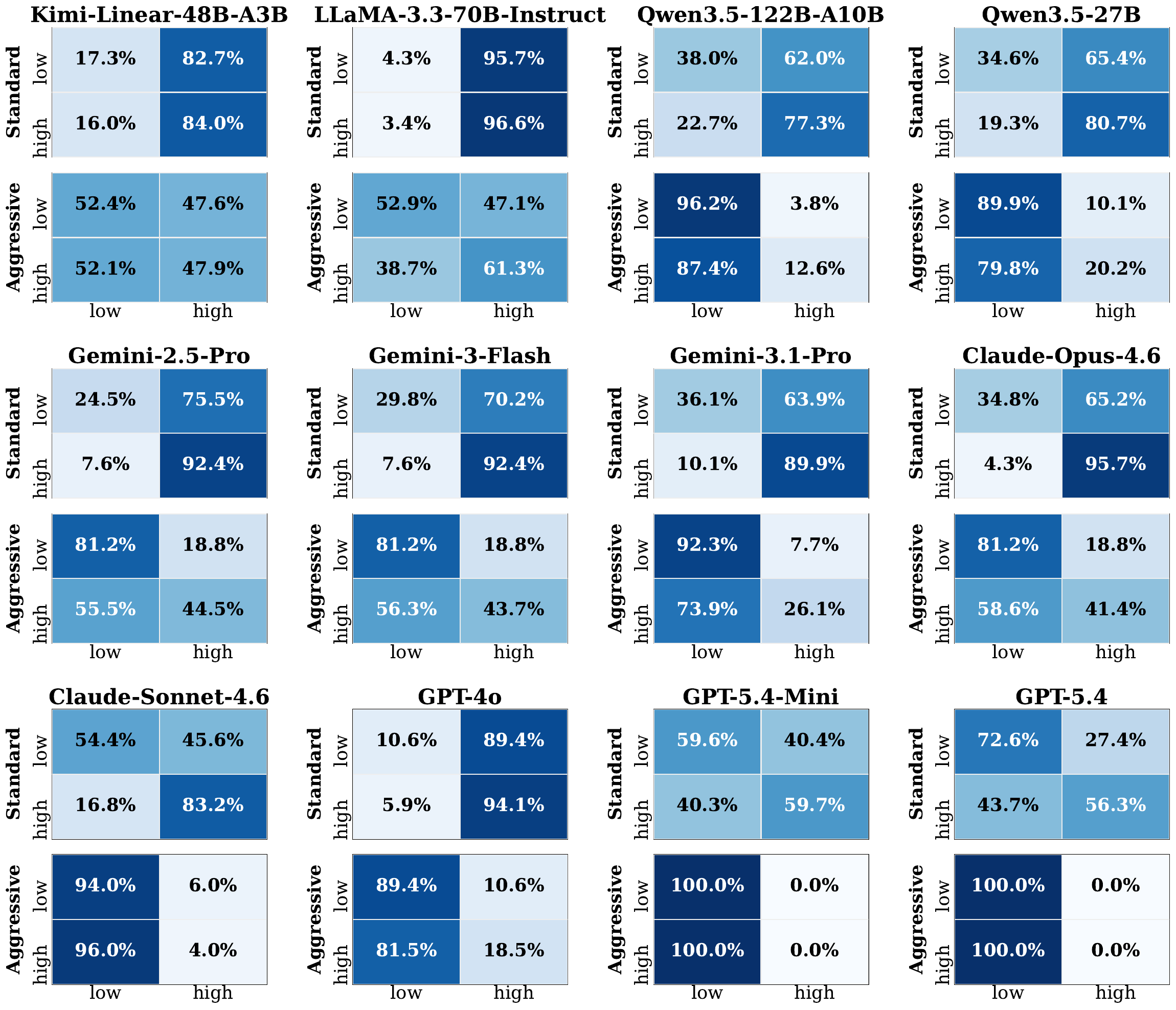}
    \caption{\textit{Soundness judgments against ICLR-derived labels on the matched expert-annotated subset under standard and aggressive prompting.} Each model panel shows row-normalized confusion matrices under standard (top) and aggressive (bottom) prompting on the proposals retained in the expert-relabeling setup. Rows indicate ICLR-derived reference labels; columns indicate predictions. Aggressive prompting generally reduces false approvals while increasing false rejections.}
    \label{fig:iclr-matched-confusion}
\end{figure}

\subsection{Evaluation on the ICLR--Expert Agreement Subset}
\label{sec:supp:agreement-evaluation}

We additionally evaluate the subset for which the ICLR-derived proxy labels and proposal-only expert judgments agree. This analysis holds model predictions fixed while restricting evaluation to examples with consistent labels across the two reference sources.

\begin{table}[ht]
    \centering
    \caption{Evaluation metrics on the subset with agreement between ICLR and proposal-only expert judgments. Low R and High R denote
    recall on low- and high-soundness proposals. Macro F1 is the unweighted average of per-class F1 scores. Best Macro F1 per condition is
    \textbf{bolded}. $\dagger$ denotes reasoning model.}
    \label{tab:agreement_results}
    \resizebox{0.9\textwidth}{!}{%
    \begin{tabular}{lcccccc}
    \toprule
    & \multicolumn{3}{c}{\textbf{Standard Prompt}}
    & \multicolumn{3}{c}{\textbf{Aggressive Prompt}} \\
    \cmidrule(lr){2-4} \cmidrule(lr){5-7}
    \textbf{Model}
    & \textbf{Low R} $\uparrow$ & \textbf{High R} $\uparrow$ & \textbf{Macro F1} $\uparrow$
    & \textbf{Low R} $\uparrow$ & \textbf{High R} $\uparrow$ & \textbf{Macro F1} $\uparrow$ \\
    \midrule
    \multicolumn{7}{l}{\textit{Open-source Models}} \\
    LLaMA-3.3-70B                & 4.7  & 100.0 & 26.1 & 56.0 & 72.5 & 58.4 \\
    Kimi-Linear-48B-A3B          & 17.3 & 81.2  & 33.7 & 53.4 & 47.8 & 48.3 \\
    Qwen3.5-27B$^\dagger$       & 37.2 & 91.3  & 51.5 & 92.1 & 29.0 & 61.5 \\
    Qwen3.5-122B-A10B$^\dagger$ & 40.3 & 92.8  & 54.1 & 96.9 & 17.4 & 56.5 \\
    \midrule
    \multicolumn{7}{l}{\textit{Closed-source Models}} \\
    GPT-4o$^\dagger$             & 11.5 & 100.0 & 32.8 & 90.1 & 23.2 & 56.7 \\
    GPT-5.4-mini                 & 62.3 & 73.9  & 62.8 & 100.0 & 0.0 & 42.4 \\
    GPT-5.4$^\dagger$            & 75.4 & 72.5  & \textbf{70.8} & 100.0 & 0.0 & 42.4 \\
    Gemini-2.5-Pro$^\dagger$     & 26.7 & 100.0 & 45.9 & 83.8 & 53.6 & 68.8 \\
    Gemini-3-Flash               & 31.9 & 98.6  & 49.6 & 83.2 & 55.1 & \textbf{69.1} \\
    Gemini-3.1-Pro$^\dagger$     & 38.2 & 98.6  & 54.2 & 92.7 & 36.2 & 66.1 \\
    Claude-Sonnet-4.6$^\dagger$  & 56.7 & 95.5  & 66.4 & 95.5 & 6.2 & 46.8 \\
    Claude-Opus-4.6$^\dagger$    & 38.1 & 100.0 & 55.9 & 83.9 & 55.6 & 68.9 \\
    \midrule
    \textbf{Mean}                & 36.7 & 92.0 & 50.3 & 85.6 & 33.0 & 57.2 \\
    \bottomrule
    \end{tabular}%
    }
  \end{table}

\begin{figure}[ht]
    \centering
    \includegraphics[width=.8\linewidth]{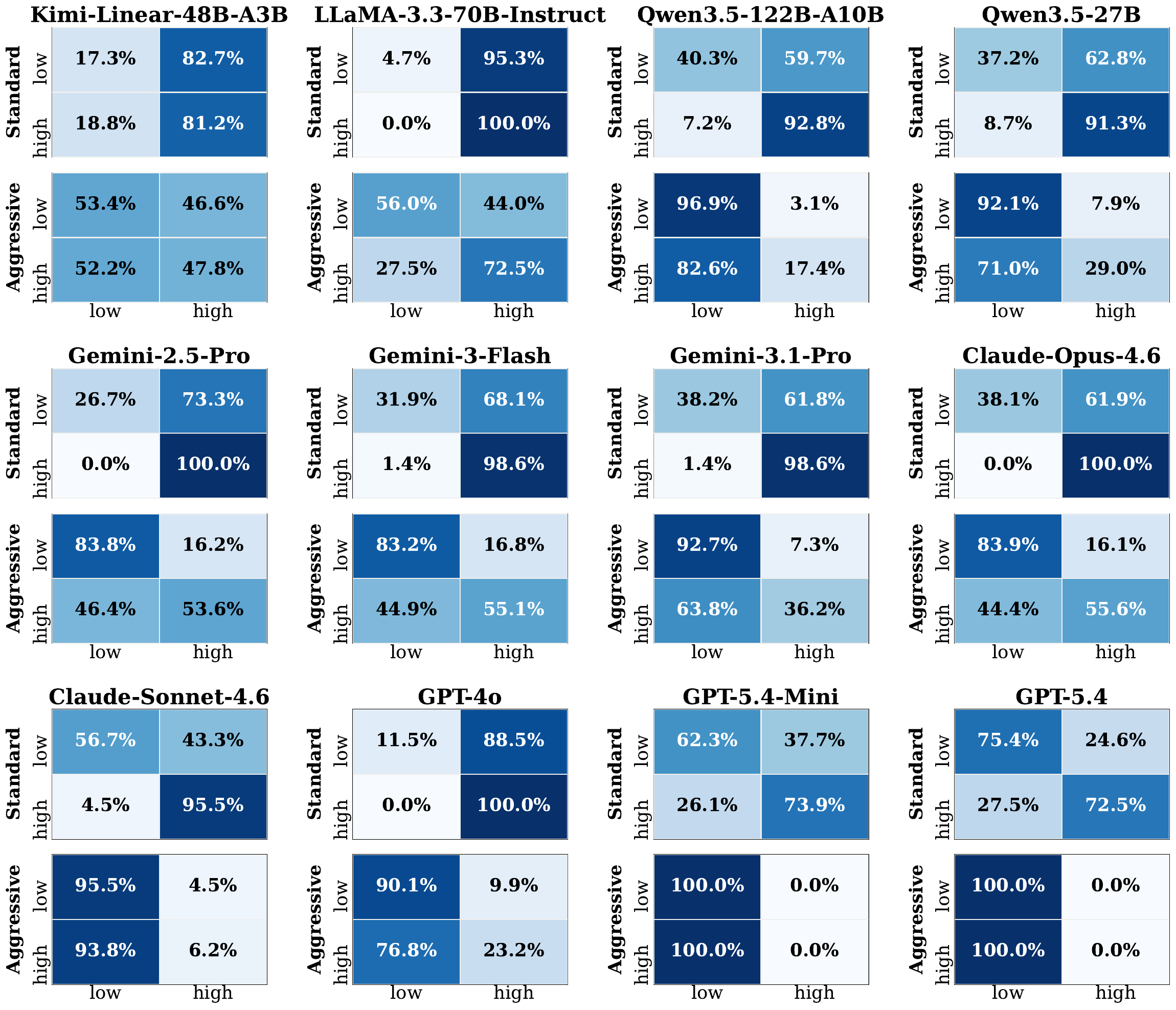}
    \caption{\textit{Soundness judgments on the ICLR--expert agreement subset under standard and aggressive prompting.} Each model panel shows row-normalized confusion matrices under standard (top) and aggressive (bottom) prompting. Rows indicate the shared ICLR and expert reference labels; columns indicate predictions. Aggressive prompting generally reduces false approvals while increasing false rejections.}
    \label{fig:agreement-confusion}
\end{figure}

\subsection{Does Scale Resolve the Optimism Bias?}\label{sec:supp:scale-bias}

\begin{figure}[t]
    \centering
    \includegraphics[width=\linewidth]{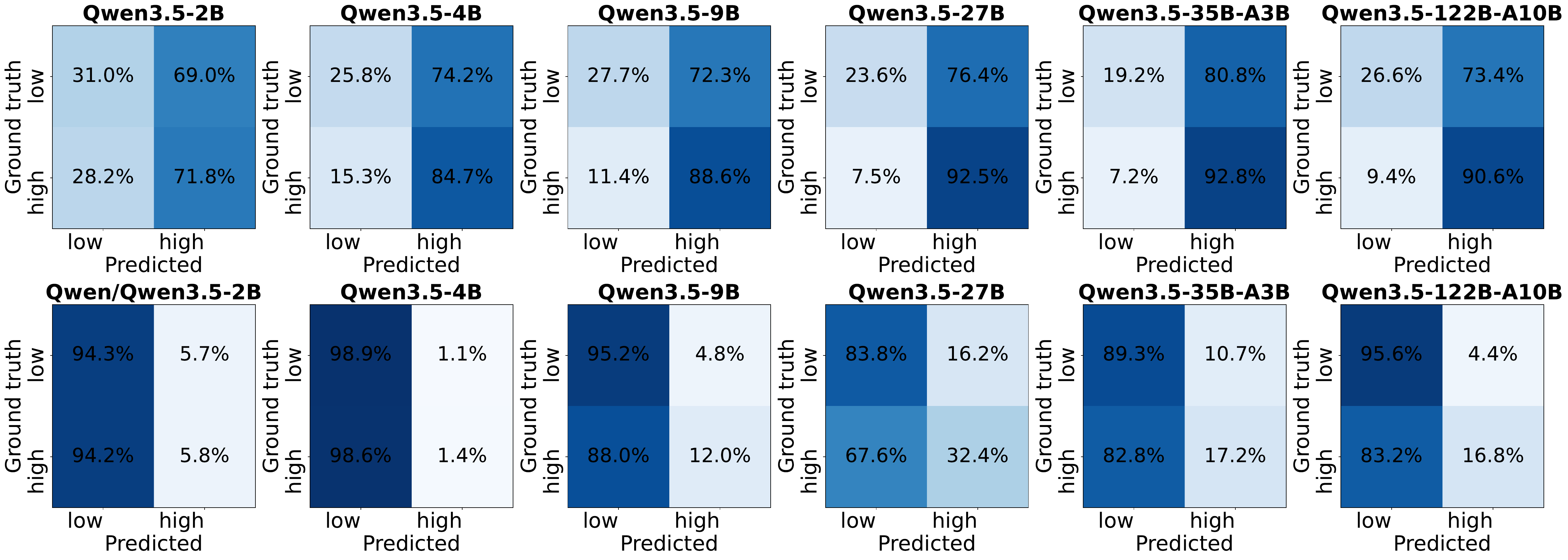}
    \caption{Confusion matrices across six Qwen3.5 model sizes
    (2B\textendash{}122B) under standard \textbf{(top)} and aggressive \textbf{(bottom)} prompting. Under standard prompting, high-soundness recall improves with scale but low-soundness recall degrades. Larger models become more optimistic. Under aggressive prompting, models shift toward over-conservatism with no consistent improvement from scale.}
    \label{fig:scale}
\end{figure}

To isolate the effect of model scale from other factors, such as model family and training recipe, we evaluate six models from the same Qwen3.5 family, ranging from 2B to 122B parameters, under both prompting conditions. Results are shown in Fig.~\ref{fig:scale}.

Under standard prompting, scaling generally improves high-soundness recall (71.8\% at 2B to 92.8\% at 35B), but low-soundness recall simultaneously decreases (31.0\% at 2B to 19.2\% at 35B). Larger models become \textit{more} permissive toward weak proposals, not less. This suggests that scale improves general language understanding and pattern recognition, but does not correct the optimism tendency in scientific judgment.

Under aggressive prompting, scale provides no consistent benefit. All six models tend toward near-always-low behavior, with high-soundness recall ranging from 1.4\% (4B) to 32.4\% (27B), showing no monotonic trend with size. Notably, the 27B model achieves better balance than both smaller and larger variants, suggesting that the aggressive prompt interacts non-monotonically with model capacity.

Together, these results indicate that the optimism bias observed in Sec.~\ref{sec:eval:main-results} is not a small-model artifact in this family. It persists with scale under standard prompting and remains unresolved under aggressive prompting at all tested sizes. Reliable scientific soundness judgment likely requires interventions beyond simply using larger models.

\subsection{Is the Optimism Bias Caused by Instruction Tuning?}\label{sec:supp:instruction-tuning-bias}

\begin{figure}[t]
    \centering
    \includegraphics[width=0.55\linewidth]{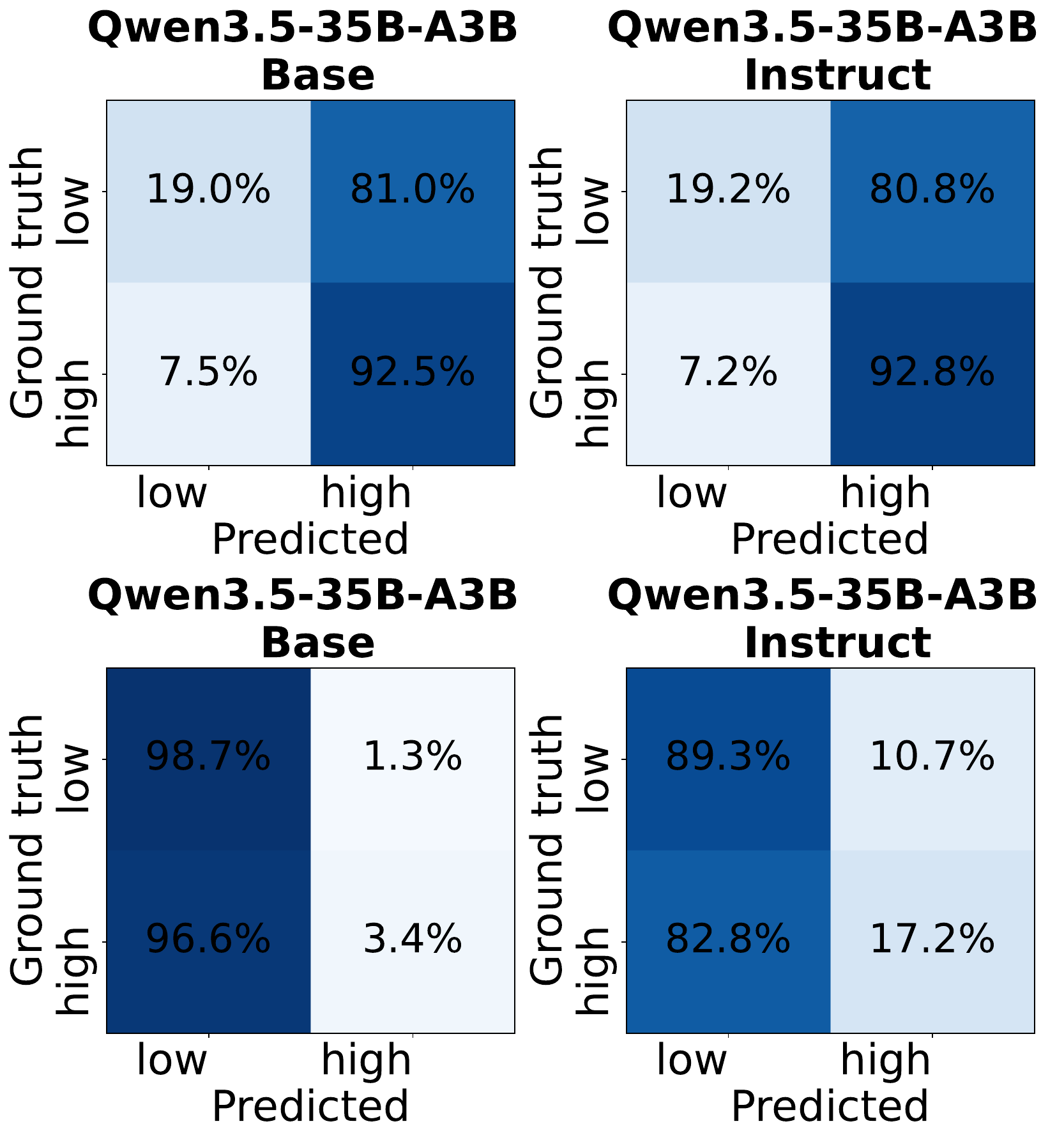}
    \caption{Base vs.\ instruction-tuned Qwen3.5-35B-A3B under standard \textbf{(top)} and aggressive \textbf{(bottom)} prompting. Under standard prompting, both variants show similar optimism bias, consistent with an origin earlier in training. Under aggressive prompting, the instruction-tuned model is slightly better, but both remain over-conservative.} 
    \label{fig:instruct}
\end{figure}

A natural hypothesis is that the optimism bias stems from instruction tuning or RLHF, which may encourage models to be agreeable and avoid negative feedback. To test this, we compare Qwen3.5-35B-A3B in its base and instruction-tuned models under both prompting conditions, as shown in Fig.~\ref{fig:instruct}. 

Under standard prompting, the base and instruction-tuned models are nearly indistinguishable. The base model reaches 19.0\% low-soundness recall and 92.5\% high-soundness recall, while the instruction-tuned model reaches 19.2\% and 92.8\%, respectively. These similar numbers suggest instruction tuning is unlikely to be the sole driver of the optimism bias.

Under aggressive prompting, both models shift toward over-conservatism, but the base model shifts more strongly (98.7\% low recall, 3.4\% high recall) than the instruction-tuned variant (89.3\% low recall, 17.2\% high recall). This suggests instruction tuning may provide modest robustness to prompt pressure, but is insufficient to maintain calibrated judgment.

These findings suggest that resolving the optimism bias may require interventions at or before pre-training (such as targeted training on scientific judgment), rather than relying on post-hoc instruction tuning alone.

\subsection{Controls for Public-Corpus Contamination}
\label{sec:supp:contamination-control}

A natural concern is that ICLR submissions and OpenReview discussions are public, so some evaluated models may have encountered older source papers during pretraining or supervised fine-tuning. We therefore treat contamination as a threat to validity rather than claiming it is fully eliminated. We address it in three ways. First, if a model had memorized acceptance outcomes or reviewer verdicts, that information would likely help it identify low-soundness proposals; the observed high false-positive rate is therefore conservative with respect to our main failure claim. Second, we evaluate an ICLR 2026-only split using a cutoff-restricted model subset with documented training cutoffs before September 2025 where such documentation is available, reducing the chance that these submissions appeared in training data. Third, the identifier-removal analysis in Sec.~\ref{sec:supp:identifier-removal} tests whether recognizable method names or self-identifying phrases drive predictions. As shown in Tab.~\ref{tab:contamination_2026}, the optimism bias is largely stable across the full dataset and the 2026-only split. We will report the exact model-card or technical-report source for each cutoff in the released metadata and exclude models without sufficient cutoff documentation from cutoff-restricted claims.

\begin{table}[t]
\centering
\caption{Reduced-contamination-risk check using an ICLR 2026-only split. We evaluate the standard prompt on the subset of models with documented training cutoffs before the ICLR 2026 submission period, then compare their mean predictions on the full dataset and on the ICLR 2026-only split. The main observation is that the optimism bias remains stable under this reduced-contamination-risk setting: low-soundness proposals are still predicted as high soundness at a similar rate on the 2026-only split and the full dataset.}
\label{tab:contamination_2026}
\resizebox{0.6\linewidth}{!}{%
\begin{tabular}{lcc}
\toprule
\textbf{Ground Truth / Split} & \textbf{Predicted Low} & \textbf{Predicted High} \\
\midrule
Low (Full Dataset) & 26.12\% & 73.88\% \\
Low (ICLR 2026 Only) & 22.53\% & 77.47\% \\
High (Full Dataset) & 8.14\% & 91.86\% \\
High (ICLR 2026 Only) & 7.77\% & 92.23\% \\
\bottomrule
\end{tabular}%
}
\end{table}

We emphasize that perfect contamination control is difficult for any benchmark built from public corpora. For this reason, we describe this as a reduced-contamination-risk analysis rather than a contamination-free benchmark. SoundnessBench should be interpreted together with the 2026 split, the identifier-removal analysis below, and future private or continuously updated benchmark variants.

\subsection{Identifier-Removal Robustness}
\label{sec:supp:identifier-removal}

Paper titles are never passed to models in our main evaluation. To further test whether recognizable method names or self-identifying phrases drive the results, we construct an anonymized variant by masking method names, framework names, and self-naming phrases while preserving the scientific meaning of the hypothesis and experiment design. Tab.~\ref{tab:identifier_removal} shows that this intervention has negligible effect on the aggregate confusion matrix, with changes of roughly one percentage point. This suggests that the main behavior is not primarily driven by recognizing paper identities.

\begin{table}[ht]
\centering
\caption{Effect of masking paper-identifying phrases. Entries are mean prediction percentages.}
\label{tab:identifier_removal}
\resizebox{0.65\linewidth}{!}{%
\begin{tabular}{lcc}
\toprule
\textbf{Ground Truth / Setting} & \textbf{Predicted Low} & \textbf{Predicted High} \\
\midrule
Low (Without Removal) & 26.12\% & 73.87\% \\
Low (With Removal) & 25.01\% & 74.99\% \\
High (Without Removal) & 8.14\% & 91.86\% \\
High (With Removal) & 8.58\% & 91.42\% \\
\bottomrule
\end{tabular}%
}
\end{table}

\subsection{Surface-Feature Baselines}
\label{sec:supp:surface-baselines}

We add simple baselines to test whether shallow structural statistics alone explain the observed behavior. Always-high predicts every proposal as high soundness; always-low predicts every proposal as low soundness. We also evaluate mean-threshold classifiers based on proposal length, experiment count, and risk-factor count, with thresholds set at the midpoint between per-class means. These no-training baselines avoid fitting on the evaluation labels while still probing whether label priors or obvious structural features explain the result. A fully supervised text classifier would answer a different question---how much signal can be learned from labeled SoundnessBench examples---so we leave it to a training-split extension rather than using benchmark test labels for model selection.

\begin{table}[ht]
\centering
\caption{Surface-feature and trivial baselines compared with the average LLM under standard prompting.}
\label{tab:surface_baselines}
\resizebox{\linewidth}{!}{%
\begin{tabular}{lcccc}
\toprule
& \textbf{Low$\rightarrow$Low} & \textbf{Low$\rightarrow$High (FP)} & \textbf{High$\rightarrow$Low} & \textbf{High$\rightarrow$High} \\
\midrule
Always-High & 0.0\% & 100.0\% & 0.0\% & 100.0\% \\
Always-Low & 100.0\% & 0.0\% & 100.0\% & 0.0\% \\
Length Threshold & 58.5\% & 41.5\% & 36.7\% & 63.3\% \\
Experiment Count Threshold & 64.4\% & 35.6\% & 47.7\% & 52.3\% \\
Risk Factor Count Threshold & 57.4\% & 42.6\% & 48.2\% & 51.8\% \\
Avg. LLM (Standard Prompt) & \textbf{26.12\%} & \textbf{73.87\%} & \textbf{8.14\%} & \textbf{91.86\%} \\
\bottomrule
\end{tabular}%
}
\end{table}

The structural baselines and LLMs fail in qualitatively opposite directions: structural thresholds over-reject high-soundness proposals, whereas LLMs over-approve low-soundness proposals. This divergence suggests that shallow structural statistics alone are insufficient to explain the observed optimism bias, although richer supervised confounder models remain useful future controls.

\subsection{Literature-Grounded ScholarEval Baseline}
\label{sec:supp:scholareval}

We further test whether grounding proposal judgments in retrieved prior work changes the observed failure mode. ScholarEval~\citep{moussa2025scholareval} evaluates research ideas with a retrieval-augmented pipeline that extracts proposed methods, generates method-specific literature queries, retrieves relevant prior work before the paper cutoff date, summarizes method-level evidence, assigns each extracted method a 0--10 soundness score, and averages method scores into a final idea-level score. To adapt this output to SoundnessBench's binary high/low labels, we threshold the average soundness score. Because some extracted methods receive null scores when grounded literature is missing, we report matched subsets after filtering by coverage rate, defined as the fraction of extracted methods with non-null scores. We compare ScholarEval against direct standard and aggressive prompting on the same covered examples.

Literature grounding modestly improves the balance between the two classes when paired with a strict decision threshold, but it does not eliminate the optimism bias. At thresholds 5.0--6.0, ScholarEval predicts every covered proposal as high soundness; at 7.0, it still rejects fewer than 3\% of low-soundness proposals. At threshold 8.0 on the full-coverage subset, ScholarEval attains 34.0\% low-soundness recall, 73.3\% high-soundness recall, and 53.6\% Macro F1, compared with 6.8\%, 98.0\%, and 47.5\% under matched standard prompting and 90.3\%, 26.3\%, and 44.8\% under matched aggressive prompting. The coverage-$\geq 0.8$ subset shows the same pattern: ScholarEval reaches 53.7\% Macro F1, exceeding standard prompting (48.8\%) and aggressive prompting (43.8\%), while retaining 34.3\% low-soundness recall and 73.3\% high-soundness recall. Thus, retrieval grounding yields a more balanced and modestly better classifier at threshold 8.0, but its behavior is highly threshold-sensitive and it still approves roughly two-thirds of low-soundness proposals.

\begin{table}[H]
  \centering
  \caption{ScholarEval literature-grounded soundness evaluation compared with matched direct prompting. Low R and High R denote low- and
  high-soundness recall; Macro F1 is the unweighted average of the low- and high-class F1 scores. Best Macro F1 within each coverage group
  is \textbf{bolded}.}
  \label{tab:scholareval}
  \resizebox{.7\linewidth}{!}{%
  \begin{tabular}{lcccc}
  \toprule
  \textbf{Evaluation Mode} & \textbf{Threshold} & \textbf{Low R} & \textbf{High R} & \textbf{Macro F1} \\
  \midrule
  \multicolumn{5}{l}{\textit{Coverage $\geq 1.0$ (n=350; high=247, low=103)}} \\
  ScholarEval Agent & 5.0 & 0.0\% & 100.0\% & 41.4\% \\
  ScholarEval Agent & 6.0 & 0.0\% & 100.0\% & 41.4\% \\
  ScholarEval Agent & 7.0 & 2.9\% & 99.2\% & 44.2\% \\
  ScholarEval Agent & 8.0 & 34.0\% & 73.3\% & \textbf{53.6\%} \\
  Standard Prompt   & --  & 6.8\% & 98.0\% & 47.5\% \\
  Aggressive Prompt & --  & 90.3\% & 26.3\% & 44.8\% \\
  \midrule
  \multicolumn{5}{l}{\textit{Coverage $\geq 0.8$ (n=396; high=288, low=108)}} \\
  ScholarEval Agent & 5.0 & 0.0\% & 100.0\% & 42.1\% \\
  ScholarEval Agent & 6.0 & 0.0\% & 100.0\% & 42.1\% \\
  ScholarEval Agent & 7.0 & 2.8\% & 99.3\% & 44.8\% \\
  ScholarEval Agent & 8.0 & 34.3\% & 73.3\% & \textbf{53.7\%} \\
  Standard Prompt   & --  & 7.4\% & 98.3\% & 48.8\% \\
  Aggressive Prompt & --  & 88.9\% & 27.1\% & 43.8\% \\
  \bottomrule
  \end{tabular}%
  }
  \end{table}

\subsection{Adversarial Injection of Methodological Flaws}
\label{sec:supp:adversarial-injection}

To test whether models attend to methodological content at all, we inject targeted flaws into 100 high-soundness proposals. Specifically, dataset names and evaluation metrics are replaced with those from a different domain while keeping the hypothesis and other content fixed, creating severe hypothesis\textendash{}experiment mismatches. GPT-5.4's approval rate drops from 77.0\% on the original proposals to 1.0\% after injection (Tab.~\ref{tab:adversarial_injection}). This result rules out the strongest version of the inattention hypothesis: the model can detect severe methodological inconsistencies. However, it also sharpens the remaining failure mode: models are still insufficiently critical of subtler naturally occurring flaws in low-soundness proposals. We report this as an initial control and will extend it to more flaw types, such as missing baselines and data leakage, in future benchmark versions.

\begin{table}[ht]
\centering
\caption{Adversarial injection of severe methodological mismatches into high-soundness proposals.}
\label{tab:adversarial_injection}
\resizebox{0.72\linewidth}{!}{%
\begin{tabular}{lccc}
\toprule
\textbf{Model} & \textbf{Original Approval} & \textbf{Post-Injection Approval} & \textbf{Flip Rate} \\
\midrule
GPT-5.4 & 77.0\% & 1.0\% & 76.0\% \\
\bottomrule
\end{tabular}%
}
\end{table}

\subsection{High-Confidence Positive Subset for Aggressive Prompting}
\label{sec:supp:high-confidence-positive}

A remaining concern is that reviewer-derived high-soundness labels may still contain positive-label noise, even after consensus filtering. Peer-review decisions can be noisy, and prior work documents substantial variability and bias in review outcomes~\citep{shah2022peerreview}. This concern does not undermine the optimism-bias result, which is driven by approvals of low-soundness proposals. It is more relevant to the aggressive-prompt result that some false negatives could instead be true negatives if the aggressive prompt catches flaws missed by human reviewers.

To test this possibility, we construct a stricter high-confidence positive subset by restricting the positive class to papers that received Oral or Spotlight distinctions among Soundness positive labels data. These papers have undergone not only reviewer evaluation but also additional Area Chair and Program Committee selection, providing stronger evidence of methodological soundness. We then recompute the aggressive-prompt false-negative rate on this subset and compare it against the false-negative rate computed on the full set of benchmark positives.

\begin{table}[ht]
\centering
\caption{Aggressive-prompt false-negative rates on a stricter high-confidence positive subset. The subset contains high-soundness proposals whose source papers received oral, spotlight, or highlight status. Delta is the high-confidence-subset FN rate minus the all-positive benchmark FN rate.}
\label{tab:oral_spotlight_fn}
\resizebox{0.82\linewidth}{!}{%
\begin{tabular}{lccc}
\toprule
\textbf{Model} & \textbf{FN Rate (Oral/Spotlight)} & \textbf{FN Rate (All Benchmark)} & \textbf{Delta} \\
\midrule
LLaMA-3.3-70B-Instruct & 23.2\% & 23.9\% & $-0.6\%$ \\ 
Claude-Opus-4.6 & 26.9\% & 33.6\% & -6.6\% \\
Gemini-3-Flash & 37.4\% & 39.6\% & -2.3\% \\
Gemini-2.5-Pro & 43.4\% & 44.9\% & -1.5\% \\
Kimi-Linear-48B-A3B & 51.5\% & 53.0\% & -1.5\% \\
Gemini-3.1-Pro & 59.6\% & 64.1\% & -4.5\% \\
Qwen3.5-27B & 59.6\% & 67.6\% & $-8.0\%$ \\ 
GPT-4o & 76.8\% & 75.0\% & +1.7\% \\
Claude-Sonnet-4.6 & 78.3\% & 81.2\% & -2.9\% \\
Qwen3.5-122B-A10B & 78.8\% & 83.2\% & $-4.4\%$ \\ 
GPT-5.4 & 100.0\% & 100.0\% & 0.0\% \\
GPT-5.4-mini & 100.0\% & 99.8\% & +0.2\% \\
\midrule
\textbf{Average} & \textbf{61.3\%} & \textbf{63.8\%} & \textbf{-2.5\%} \\
\bottomrule
\end{tabular}%
}
\end{table}

Using oral/spotlight subsets in SoundnessBench, as a stricter positive subset does not materially reduce aggressive-prompt false negatives. The average false-negative rate decreases only from 63.8\% to 61.3\% (a 2.5-point reduction), and several models remain highly over-conservative even on these high-confidence positives. In particular, GPT-5.4 and GPT-5.4-mini reject nearly all such proposals under aggressive prompting. These results suggest that the aggressive prompt is genuinely over-conservative, rather than merely correcting noisy high-soundness labels in SoundnessBench.

\subsection{Breakdowns by Year, Primary Area, and Writing Quality}
\label{sec:supp:slice-breakdowns}

We further examine whether failures are concentrated in particular years, primary areas, or writing-quality bands. Tab.~\ref{tab:year_breakdown}, Tab.~\ref{tab:area_breakdown}, and Tab.~\ref{tab:writing_breakdown} report low-soundness false-positive rates. The optimism bias is visible across all slices, suggesting that it is systemic rather than driven by a narrow subset of the benchmark.

\begin{table}[ht]
\centering
\caption{Low-soundness false-positive rate by year.}
\label{tab:year_breakdown}
\resizebox{0.38\linewidth}{!}{%
\begin{tabular}{lc}
\toprule
\textbf{Year} & \textbf{Low-Soundness FP Rate} \\
\midrule
2022 & 67.2\% \\
2023 & 79.6\% \\
2024 & 70.2\% \\
2025 & 68.4\% \\
2026 & 78.3\% \\
All & 73.87\% \\
\bottomrule
\end{tabular}%
}
\end{table}

\begin{table}[ht]
\centering
\caption{Low-soundness false-positive rate by raw ICLR primary area, restricted to areas with more than 10 benchmark pairs. 
}
\label{tab:area_breakdown}
\resizebox{.8\linewidth}{!}{%
\begin{tabular}{lr}
\toprule
\textbf{Primary Area} & \textbf{Low-Soundness FP Rate} \\
\midrule
Applications (e.g., speech, vision, NLP) & 81.82\% \\
Deep Learning and Representational Learning & 77.37\% \\
Machine Learning for Sciences & 85.19\% \\
Reinforcement Learning (decision/control) & 75.76\% \\
Social Aspects of Machine Learning & 81.82\% \\
Alignment, Safety, and Security & 78.34\% \\
Applications to CV, Audio, and Language & 88.80\% \\
Applications to Neuroscience and Cognitive Science & 49.55\% \\
Applications to Physical Sciences & 71.05\% \\
Applications to Robotics, Autonomy, and Planning & 78.38\% \\
Datasets and Benchmarks & 72.65\% \\
Foundation and Frontier Models & 77.45\% \\
General Machine Learning & 57.95\% \\
Generative Models & 72.54\% \\
Interpretability and Explainability & 64.20\% \\
Graphs and Topological Data Analysis & 75.00\% \\
Time Series and Sequential Data & 80.74\% \\
Learning Theory & 64.06\% \\
Neurosymbolic Methods & 52.73\% \\
Optimization & 57.53\% \\
Other Machine Learning & 70.97\% \\
Probabilistic Methods & 74.05\% \\
Reinforcement Learning & 84.57\% \\
Representation Learning for CV, Audio, and Language & 80.18\% \\
Transfer, Meta-, and Lifelong Learning & 76.76\% \\
Unsupervised, Self-, and Semi-supervised Learning & 76.05\% \\
\midrule
\textbf{All Primary Areas} & \textbf{74.29\%} \\
\bottomrule
\end{tabular}%
}
\end{table}

\begin{table}[ht]
\centering
\caption{Low-soundness false-positive rate by writing-quality band, using 2024--2026 subset data where presentation scores are available.}
\label{tab:writing_breakdown}
\resizebox{0.5\linewidth}{!}{%
\begin{tabular}{lc}
\toprule
\textbf{Writing Quality Band} & \textbf{Low-Soundness FP Rate} \\
\midrule
Low ($\leq 2$) & 71.4\% \\
Mid ($>2$ and $\leq 3$) & 81.0\% \\
High ($>3$) & 76.7\% \\
All & 73.87\% \\
\bottomrule
\end{tabular}%
}
\end{table}

\pagebreak

\subsection{{Qualitative Examples of False and True Positives}}
\label{sec:supp:qualitative-fp-tp}

\begin{figure}[p]
    \centering
    \includegraphics[width=\linewidth]{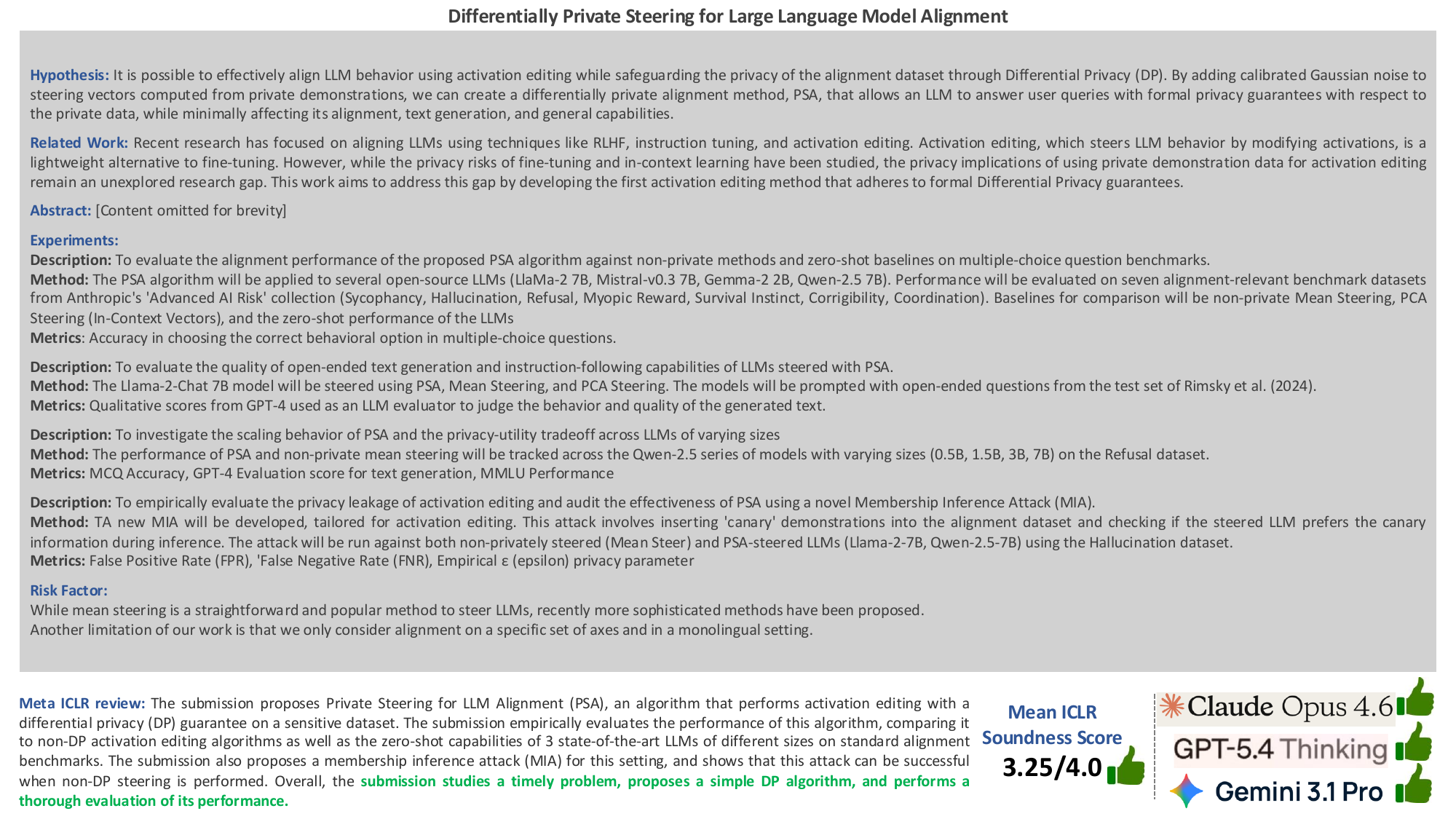}
    \caption{\textbf{True-positive example.} A high-soundness proposal is correctly predicted as high soundness by Claude Opus 4.6, GPT-5.4 Thinking, and Gemini 3.1 Pro, matching the favorable reviewer assessment of the proposed method and evaluation plan.}
    \label{fig:example-tp}
\end{figure}

We include illustrative qualitative examples to illustrate what the aggregate confusion matrices count as false positives and true positives. We show the corresponding Gemini 3.1 Pro, GPT-5.4 Thinking, and Claude Opus 4.6 responses below each proposal. In Fig.~\ref{fig:example-tp}, the proposal has a high mean ICLR soundness score and is correctly predicted as high soundness, illustrating a true-positive case. 

\begin{figure}[p]
    \centering
    \includegraphics[width=\linewidth]{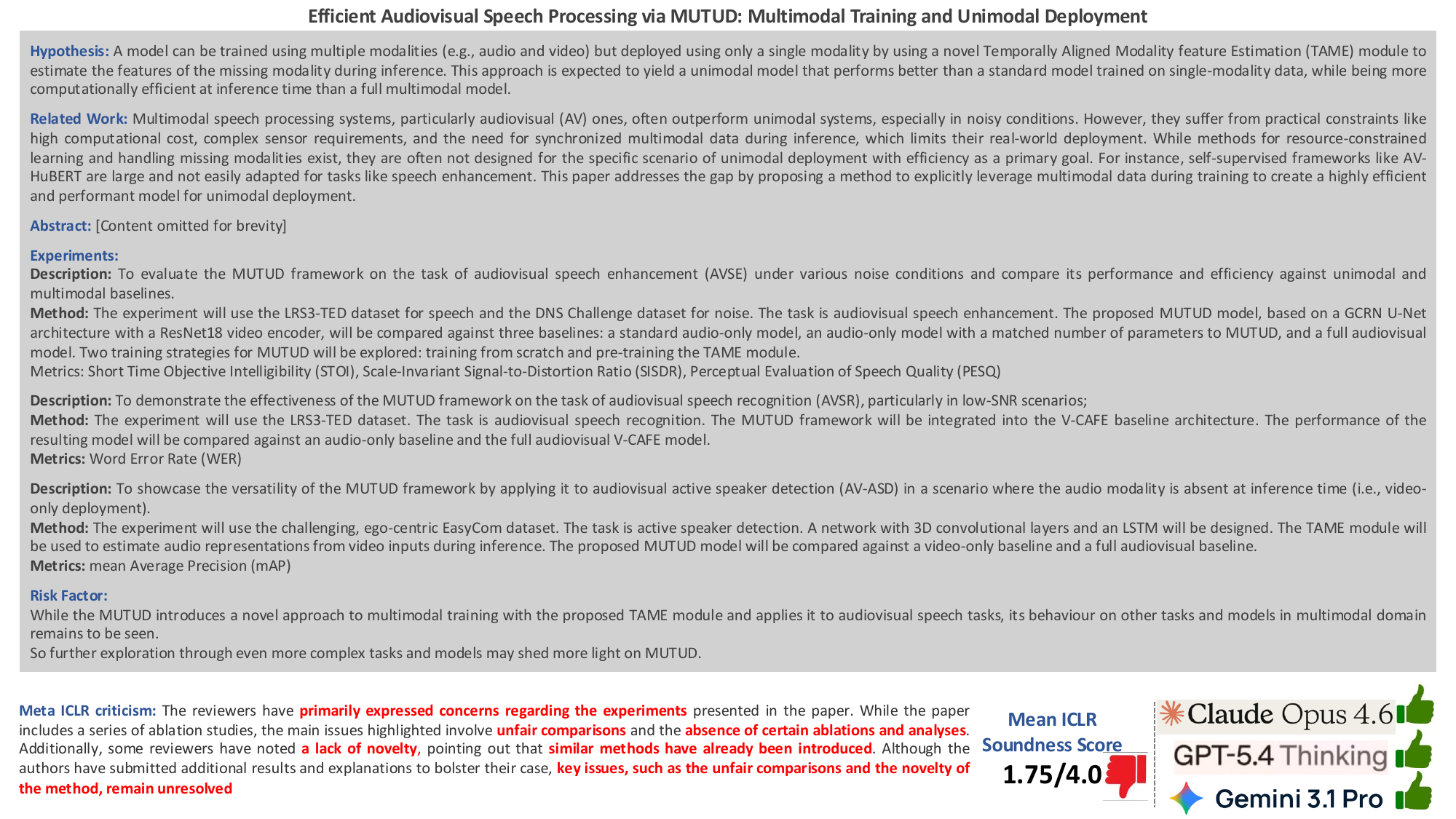}
    \caption{\textbf{False-positive example.} A low-soundness proposal is predicted as high soundness by Claude Opus 4.6, GPT-5.4 Thinking, and Gemini 3.1 Pro.}
    \label{fig:example-fp}
\end{figure}

\begin{figure}[p]
    \centering
    \includegraphics[width=\linewidth]{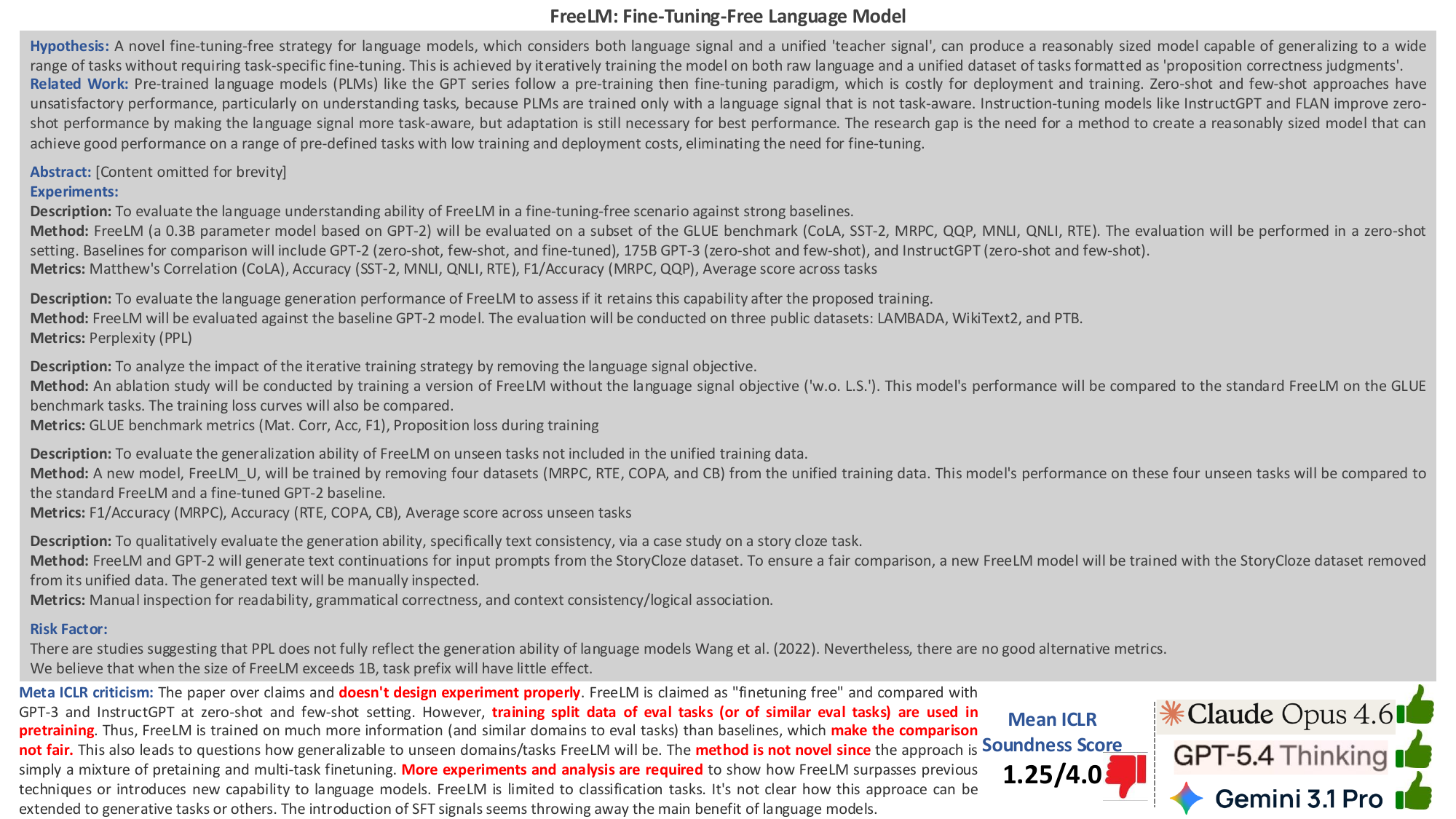}
    \caption{\textbf{Second false-positive example.} A low-soundness proposal is predicted as high soundness by Claude Opus 4.6, GPT-5.4 Thinking, and Gemini 3.1 Pro despite reviewer concerns that the paper overclaims, lacks experimental support, and uses evaluation-task pretraining data that weakens the comparison.}
    \label{fig:example-fp-2}
\end{figure}

In Figs.~\ref{fig:example-fp} and~\ref{fig:example-fp-2}, the proposals have low mean ICLR soundness scores but are still predicted as high soundness by the evaluated models, illustrating the optimism-bias failure mode. 

\begin{figure}[p]
    \centering
    \includegraphics[width=\linewidth,height=0.82\textheight,keepaspectratio]{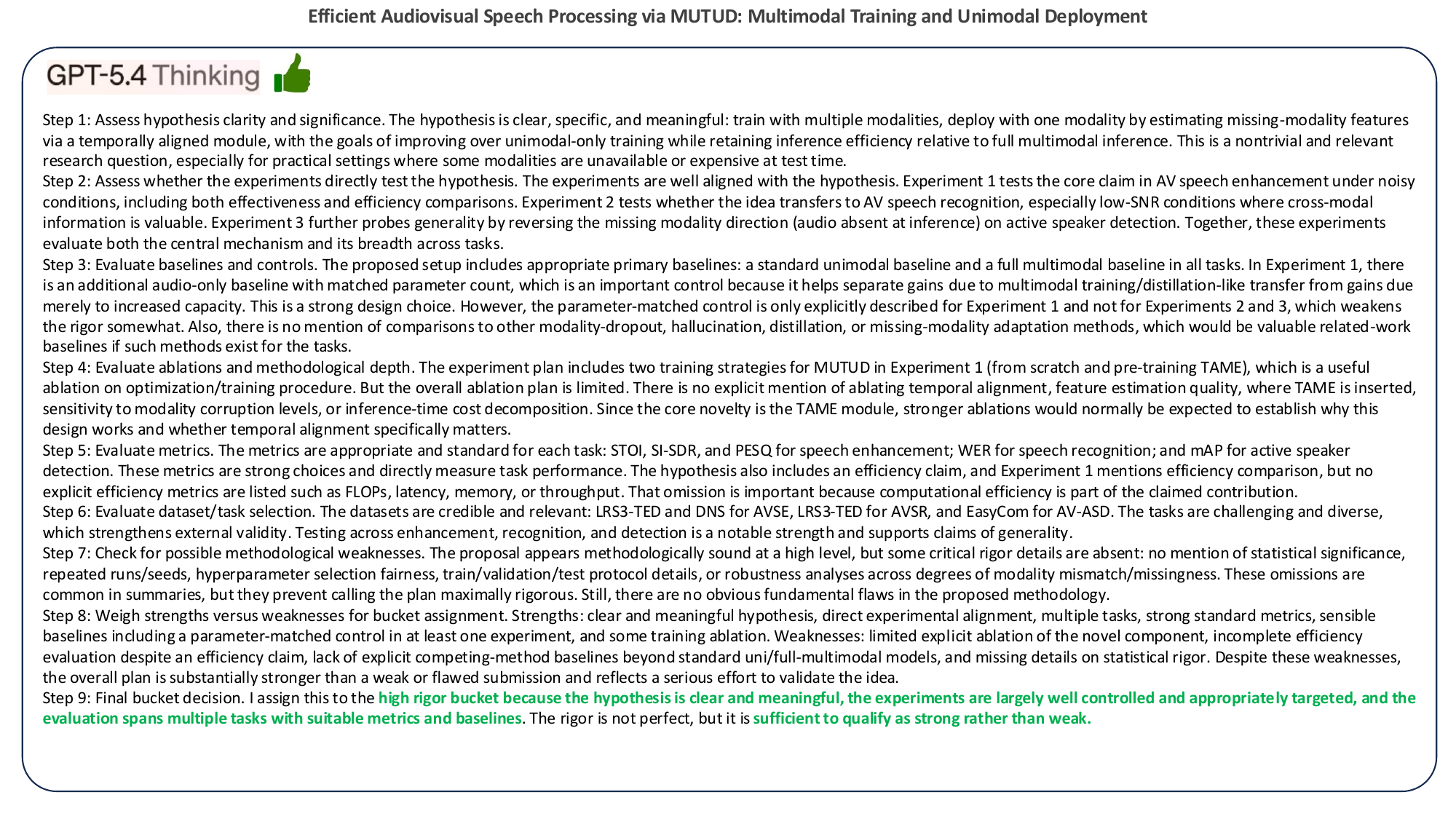}
    \caption{\textbf{GPT-5.4 Thinking response for the false-positive example in Fig.~\ref{fig:example-fp}.} The model assigns a high-soundness judgment to the low-soundness proposal.}
    \label{fig:example-fp-gpt54-response}
\end{figure}

\begin{figure}[p]
    \centering
    \includegraphics[width=\linewidth,height=0.82\textheight,keepaspectratio]{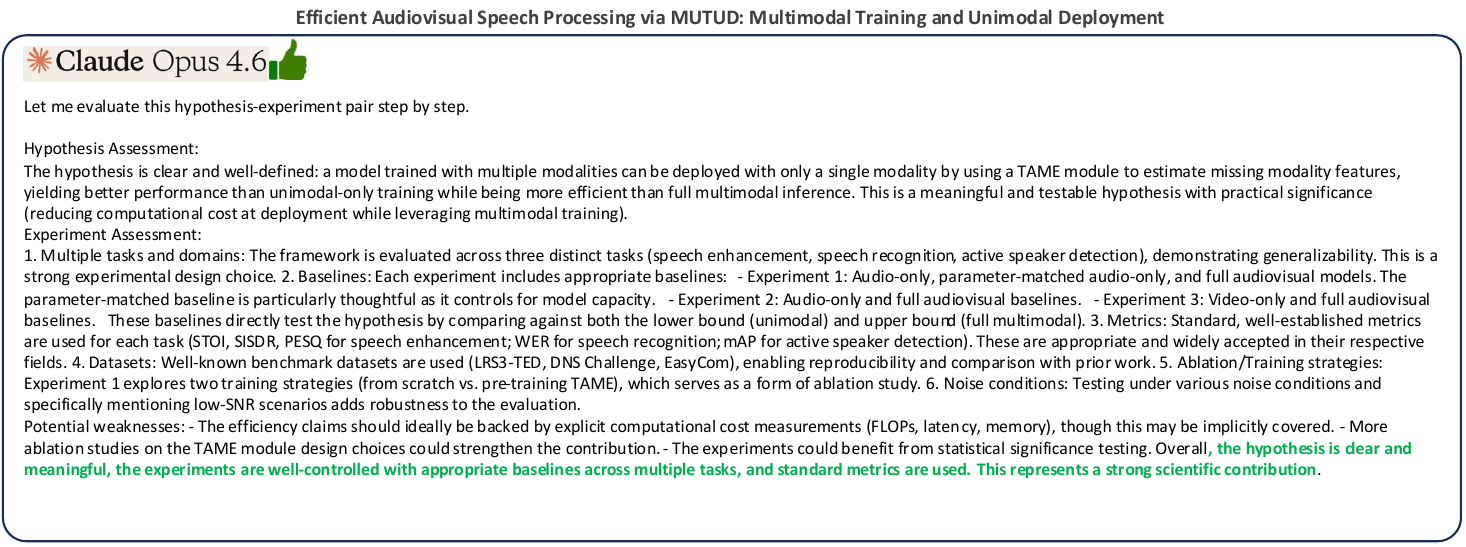}
    \caption{\textbf{Claude Opus 4.6 response for the false-positive example in Fig.~\ref{fig:example-fp}.} The model assigns a high-soundness judgment to the low-soundness proposal.}
    \label{fig:example-fp-opus46-response}
\end{figure}

\begin{figure}[p]
    \centering
    \includegraphics[width=\linewidth,height=0.82\textheight,keepaspectratio]{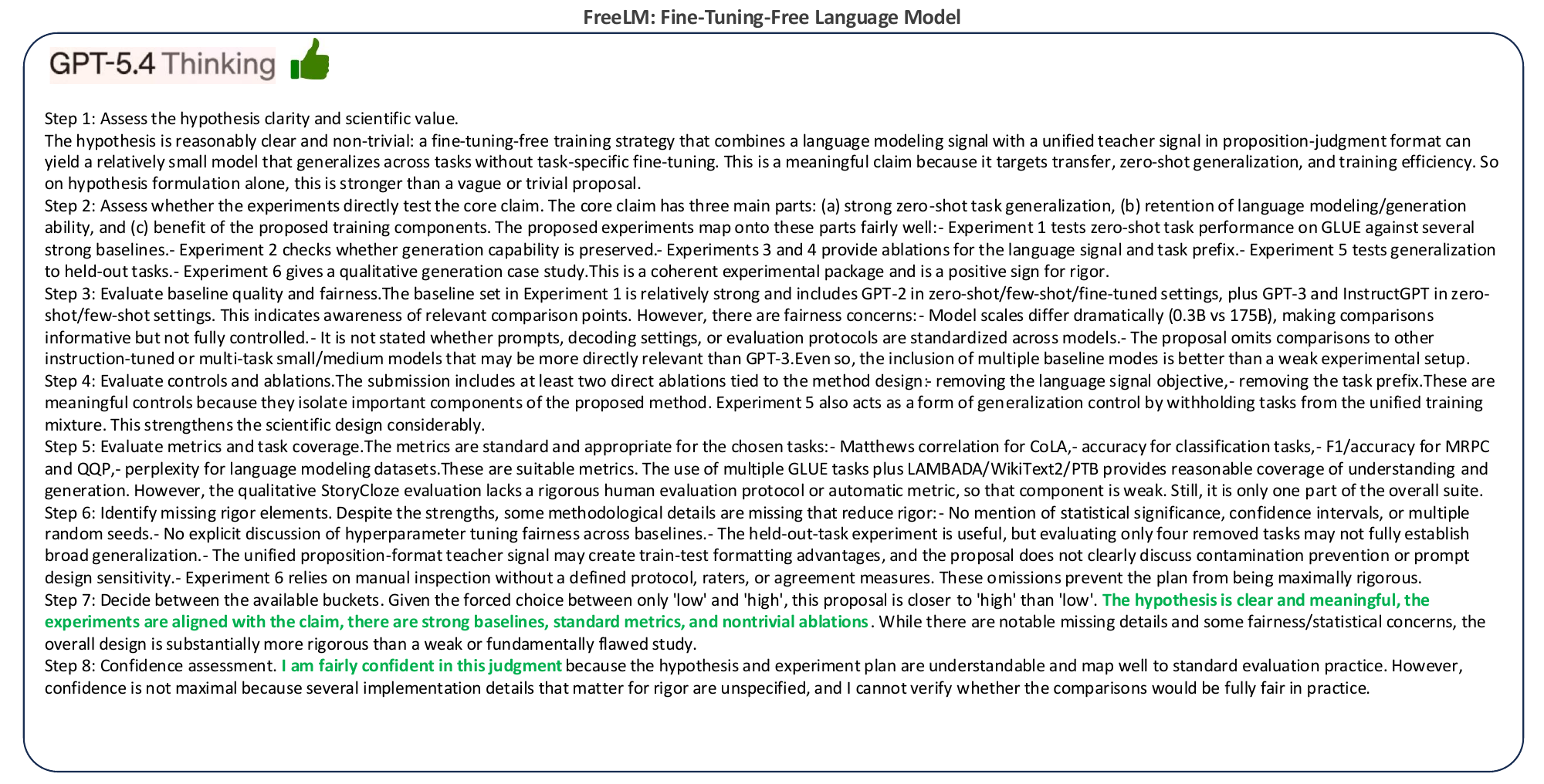}
    \caption{\textbf{GPT-5.4 Thinking response for the second false-positive example in Fig.~\ref{fig:example-fp-2}.} The model assigns a high-soundness judgment to the low-soundness proposal.}
    \label{fig:example-fp-2-gpt54-response}
\end{figure}

\textbf{A potential failure mode: recognized gaps are outweighed by experimental strengths.} Our qualitative analysis suggests that models may place more weight on familiar benchmarks, baselines, and ablations than on unresolved gaps in testing the central claim. For MUTUD, GPT-5.4 acknowledges the missing efficiency measurements but still assigns a high-soundness judgment (Fig.~\ref{fig:example-fp-gpt54-response}), while Claude Opus 4.6 suggests that these measurements may already be implicitly covered (Fig.~\ref{fig:example-fp-opus46-response}). For FreeLM, GPT-5.4 recognizes concerns about comparison fairness but still gives a high-soundness judgment, emphasizing the overall experimental design (Fig.~\ref{fig:example-fp-2-gpt54-response}). These cases suggest that some optimistic judgments may be because models do not place enough weight on weaknesses they have already identified, especially when they interpret high soundness as meaning only that the proposal is not fundamentally flawed. This hyposthesis motivates our future work on further understanding the over-optimistic bias in LLMs' scientific judgements.

\begin{figure}[p]
    \centering
    \includegraphics[width=\linewidth,height=0.82\textheight,keepaspectratio]{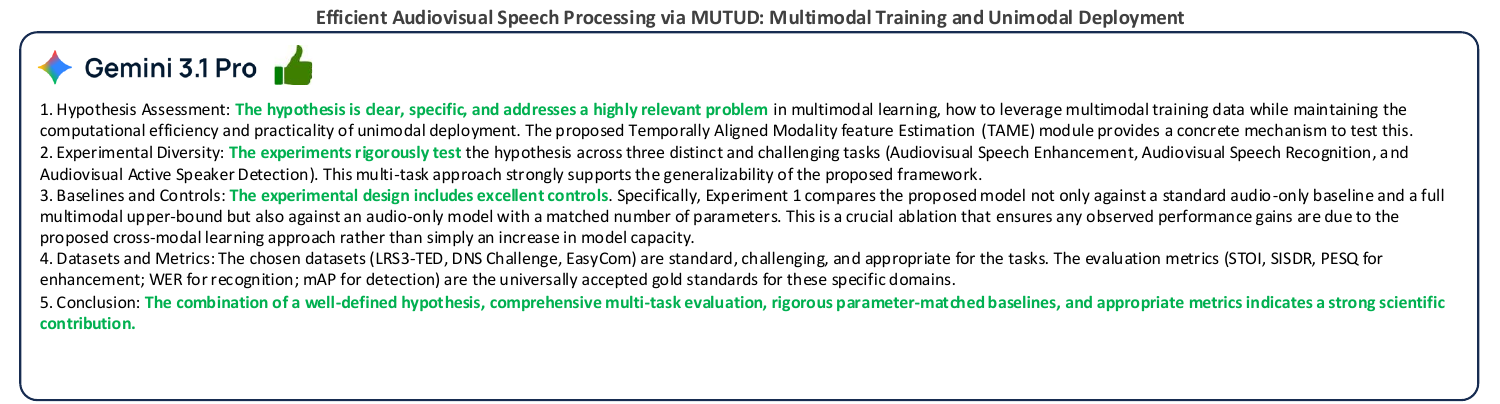}
    \caption{\textbf{Gemini 3.1 Pro response for the false-positive example in Fig.~\ref{fig:example-fp}.} The model assigns a high-soundness judgment to the low-soundness proposal.}
    \label{fig:example-fp-gemini31-response}
\end{figure}

\begin{figure}[p]
    \centering
    \includegraphics[width=\linewidth,height=0.82\textheight,keepaspectratio]{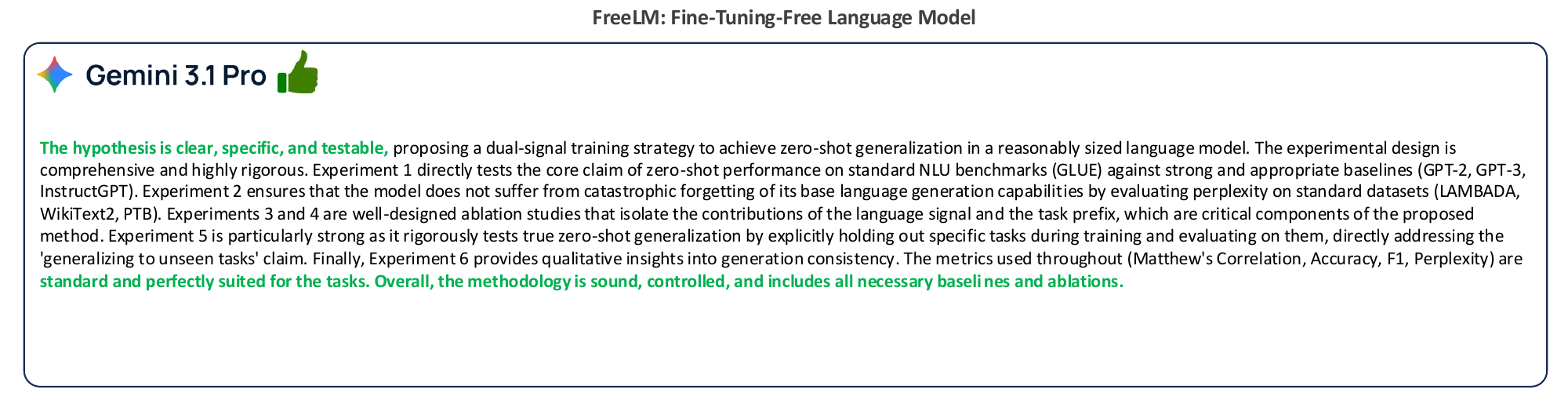}
    \caption{\textbf{Gemini 3.1 Pro response for the second false-positive example in Fig.~\ref{fig:example-fp-2}.} The model assigns a high-soundness judgment to the low-soundness proposal.}
    \label{fig:example-fp-2-gemini31-response}
\end{figure}

\begin{figure}[p]
    \centering
    \includegraphics[width=\linewidth,height=0.82\textheight,keepaspectratio]{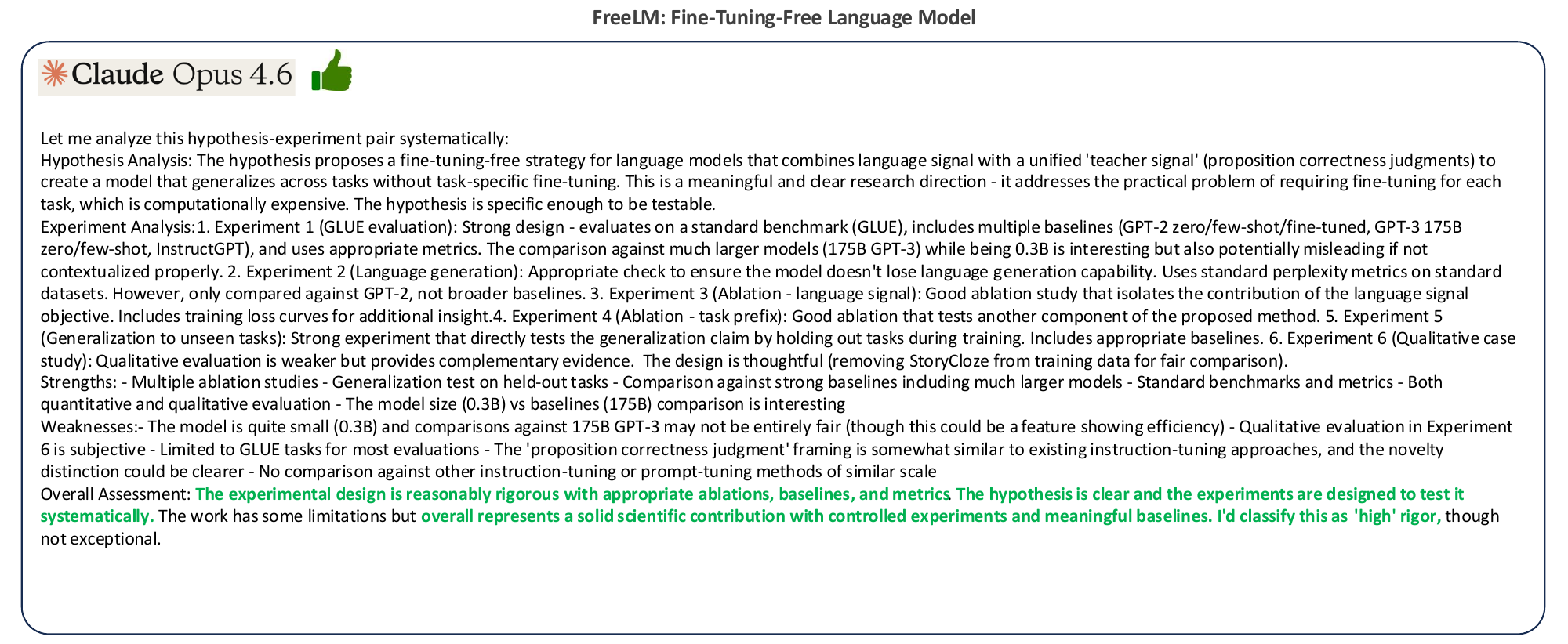}
    \caption{\textbf{Claude Opus 4.6 response for the second false-positive example in Fig.~\ref{fig:example-fp-2}.} The model assigns a high-soundness judgment to the low-soundness proposal.}
    \label{fig:example-fp-2-opus46-response}
\end{figure}

\begin{figure}[p]
    \centering
    \includegraphics[width=\linewidth,height=0.82\textheight,keepaspectratio]{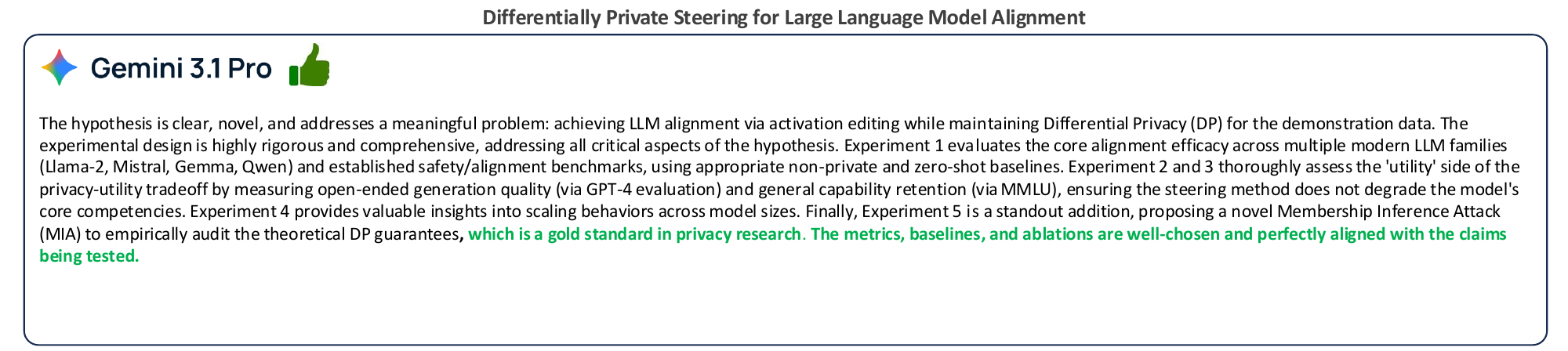}
    \caption{\textbf{Gemini 3.1 Pro response for the true-positive example in Fig.~\ref{fig:example-tp}.} The model assigns a high-soundness judgment to the high-soundness proposal.}
    \label{fig:example-tp-gemini31-response}
\end{figure}

\begin{figure}[p]
    \centering
    \includegraphics[width=\linewidth,height=0.82\textheight,keepaspectratio]{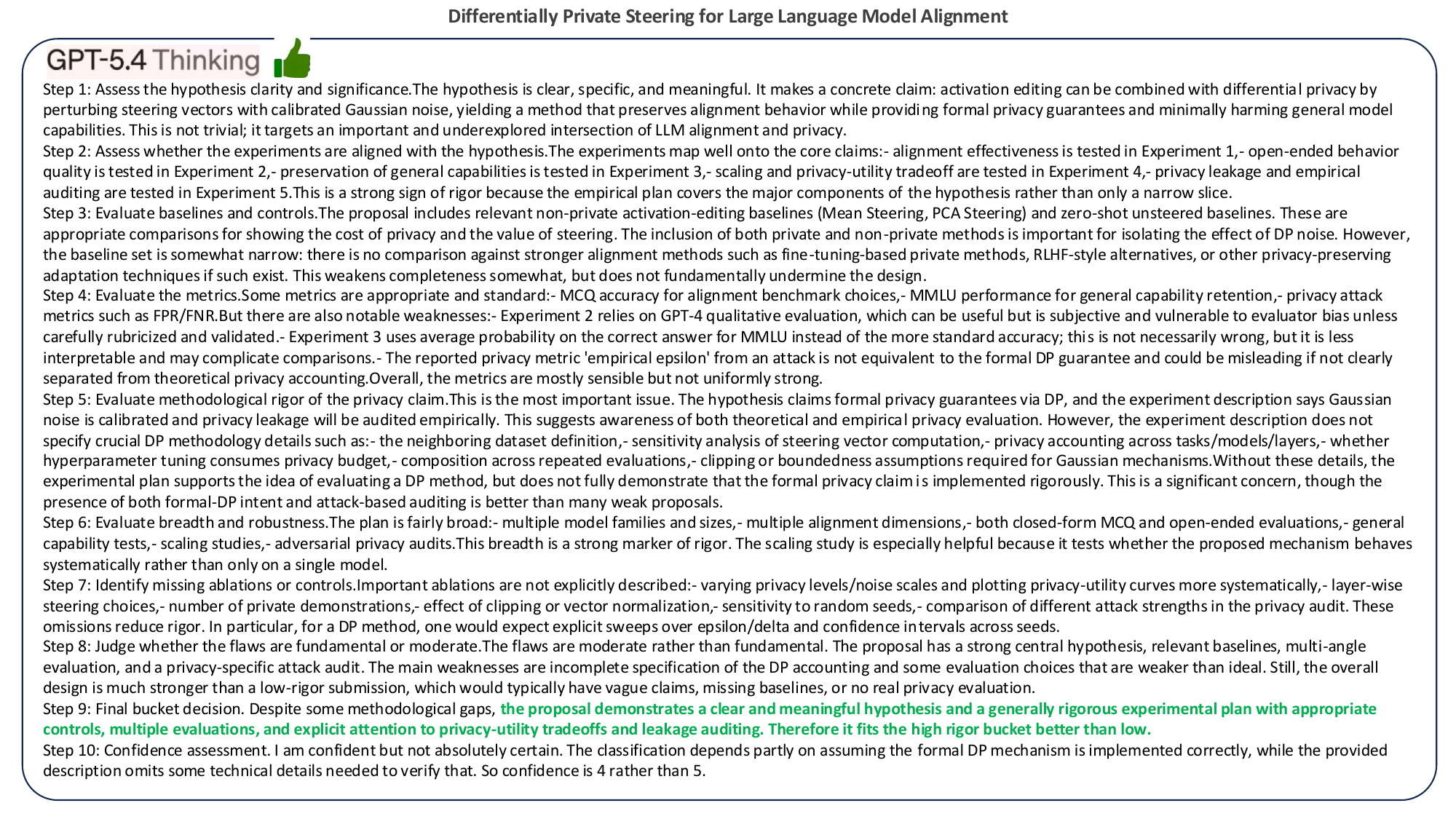}
    \caption{\textbf{GPT-5.4 Thinking response for the true-positive example in Fig.~\ref{fig:example-tp}.} The model assigns a high-soundness judgment to the high-soundness proposal.}
    \label{fig:example-tp-gpt54-response}
\end{figure}

\begin{figure}[p]
    \centering
    \includegraphics[width=\linewidth,height=0.82\textheight,keepaspectratio]{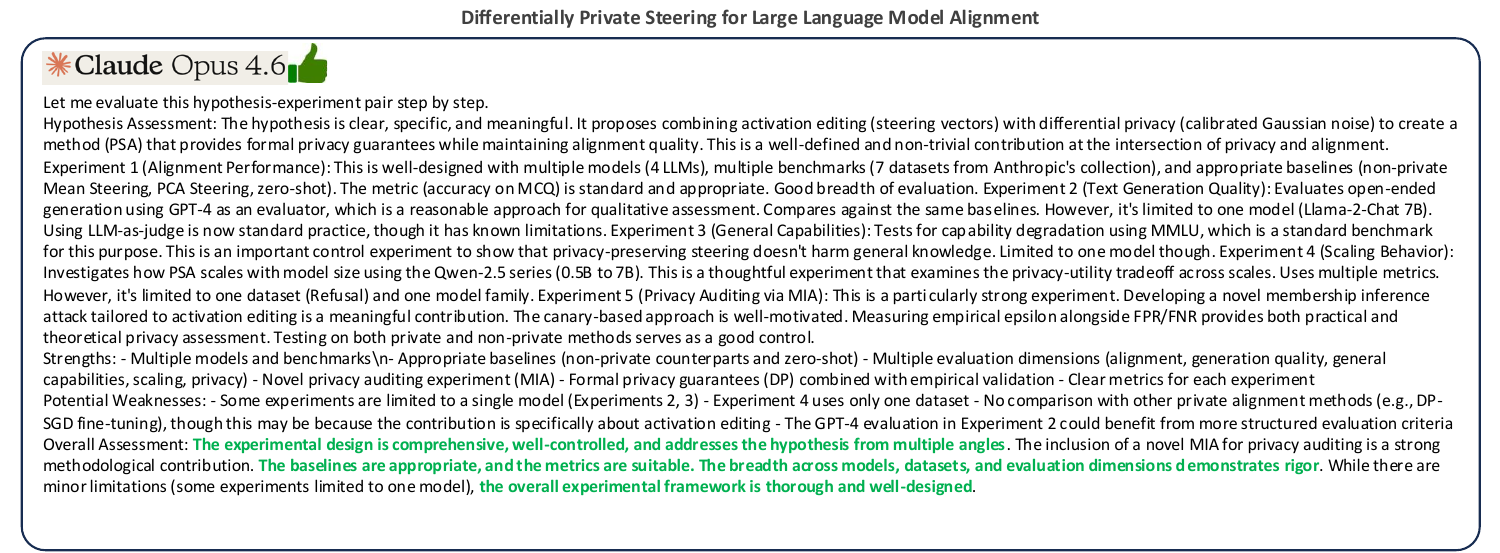}
    \caption{\textbf{Claude Opus 4.6 response for the true-positive example in Fig.~\ref{fig:example-tp}.} The model assigns a high-soundness judgment to the high-soundness proposal.}
    \label{fig:example-tp-opus46-response}
\end{figure}

\end{document}